\documentclass[sn-mathphys]{sn-jnl}

\usepackage[justification=centering]{caption}
\usepackage{extarrows}
\usepackage{multirow}
\usepackage{subfig}
\usepackage{graphicx}
\graphicspath{{./pdf/}{./jpeg/}{./png/}{./jpg/}}
\DeclareGraphicsExtensions{.pdf,.jpeg,.png,.jpg}

\jyear{2021}%

\theoremstyle{thmstyleone}%
\theoremstyle{thmstyletwo}%

\theoremstyle{thmstylethree}%

\begin{document}

\title[Article Title]{Style Generation in Robot Calligraphy with Deep Generative Adversarial Networks}



\author[1]{\fnm{Xiaoming} \sur{Wang}}\email{xiaoming.wang@um.edu.mo}
\author*[2]{\fnm{Zhiguo} \sur{Gong}}\email{fstzgg@um.edu.mo}
\affil[1]{\orgdiv{Department of Computer and Information Science}, \orgname{University of Macau}, \orgaddress{ \city{Macau}, \postcode{999078}, \country{China}}}








\abstract{Robot calligraphy is an emerging exploration of artificial intelligence in the fields of art and education. Traditional calligraphy generation researches mainly focus on methods such as tool-based image processing, generative models, and style transfer. Unlike the English alphabet, the number of Chinese characters is tens of thousands, which leads to difficulties in the generation of a style consistent Chinese calligraphic font with over 6000 characters. Due to the lack of high-quality data sets, formal definitions of calligraphy knowledge, and scientific art evaluation methods, The results generated are frequently of low quality and falls short of professional-level requirements. To address the above problem, this paper proposes an automatic calligraphy generation model based on deep generative adversarial networks (deepGAN) that can generate style calligraphy fonts with professional standards. The key highlights of the proposed method include: (1) The datasets use a high-precision calligraphy synthesis method to ensure its high quality and sufficient quantity; (2) Professional calligraphers are invited to conduct a series of Turing tests to evaluate the gap between model generation results and human artistic level; (3) Experimental results indicate that the proposed model is the state-of-the-art among current calligraphy generation methods. The Turing tests and similarity evaluations validate the effectiveness of the proposed method.}




\maketitle




\section{Introduction}\label{sec1}
Robot calligraphy serves as a visual representation of artificial intelligence applications in the arts and education. Traditional calligraphy generation researches mainly focus on the tool-based image processing \cite{kobayashi2022generative, yang2014feature, wong2008model}, generative models \cite{chen2018gated, wang2020deep}, style transfer \cite{liang2020robot}, computer graphics \cite{jian2019learning}, human gesture \cite{chao2017robot, chao2018use}, robotic motion \cite{ma2016aesthetics, wu2023internal}, etc. The vigorous development and widespread application of machine learning and deep learning in recent years have promoted research on calligraphy generation \cite{gao2019data}. However, due to the lack of high-quality datasets and the complexity of high-precision control algorithms, automatically generated calligraphy often suffers from low-quality problems and cannot Competing with human writing skills \cite{GUO2022Calligraphy, zhang2019survey}. 

Robotic calligraphy is heavily dependent on the calligraphy font datasets. Chinese character system has over 9000 commonly used characters. Even a skilled font designer finds creating a family of Chinese fonts challenging. Naturally, creating a Chinese calligraphic font is considerably more challenging, and it typically needs to take one year of manual adjusting. Whether is it possible to develop an efficient method for supporting automatic generation of Chinese calligraphic fonts? To address to this challenge problem computer scientists have been working on it for decades \cite{DBLP:journals/expert/XuJJLP09, DBLP:journals/expert/DongXZGP08}. The biggest obstacle is determining whether the algorithm can reproduce the same artistic styles from masterpieces of ancient calligraphy, including those of the most well-known ancient artworks. 

There are mainly two categories in which the prior research on automatic generation of Chinese calligraphy: calligraphy synthesizing and deep learning methods. Calligraphy synthesizing is to synthesize Chinese charters based on the designed expert systems with calligraphic knowledge representation on the given calligraphic strokes and structural style characteristics \cite{DBLP:journals/expert/XuJJLP09, DBLP:journals/expert/DongXZGP08, Zong:2014:SAP:2892753.2892971}. And calligraphy generation based on deep learning is to take characters as images for topography transformation \cite{DBLP:journals/corr/ChangG17,Lian:2016:AGL:3005358.3005371, Auto-En17}. However, the results of the previous researches are not compatible with the original characters, or the outputs are relatively handwriting fonts rather calligraphic works. Our method is based on deep learning with suitable thresholds as well additional post-processing to the generated testing characters. After a series of Turing tests and evaluations for original style fonts mixed with the generated characters, the testers can hardly recognize the generated characters from the test results. There are nearly 80\% characters are recognizable as the same artistic calligraphic style, while only few characters are not acceptable for discontinuous or mixed strokes.

This paper explores the feasibility of generating Chinese calligraphic fonts automatically with deep learning based on generative adversarial networks. We use a famous calligraphic font: Qigong Font as a case study. The result is satisfiable and applicable for practical font development, which is confirmed by two advanced font designers of FounderType Company in Beijing after Turing test. 

The innovation points of this paper are summarized as follows:

(1) The data set used in this paper comes from the calligraphy synthesis method proposed in previous work \cite{wang2023generative}. This method can synthesize the required calligraphy fonts to the greatest extent, making the data set have higher accuracy and sufficient data volume. High-quality dataset is an essential assurance for the validity of experimental findings.

(2) The proposed deep generative adversarial networks (deepGAN) is based on the GAN model, which is the state-of-the-art among current calligraphy generation methods. Case studies and result evaluation confirm the advancement of the proposed method.

(3) The Turing test is used as one of the evaluation metrics in addition to the image similarity evaluation approach. Randomly arranged calligraphers are invited to conduct the test on the experimental results to conform to the actual situation of human artistic aesthetics. 

The remaining sections of this paper are organized as follows. The approach of general deep learning algorithms is briefly discussed in Section 2, while dataset preparation is covered in Section 3. The Turing test is included in Section 4 along with a detailed analysis and discussion of the outcomes and comparisons from our experiment. Section 5 brings the paper's contribution to a conclusion. 

\section{Methodology}\label{sec2}

\begin{figure}[!htbp]
		\centering
		\includegraphics[width=1\textwidth]{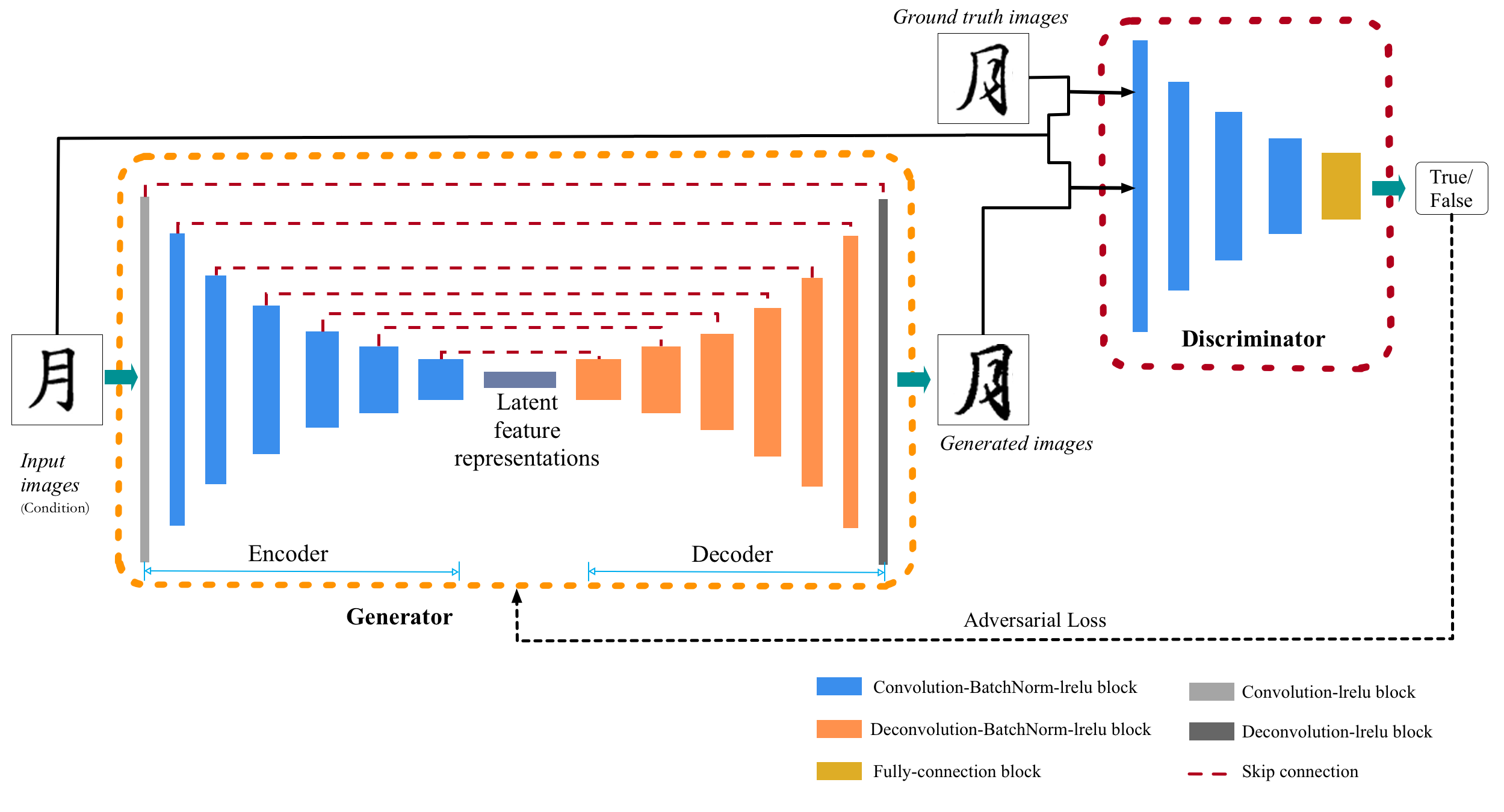}
		\caption{Overview of the proposed model.}
		\label{fig: structureofnetwork}
	\end{figure}

In this section, we intend to utilize deep learning to transfer calligraphy style from standard Kai font to QiGong font. Our approach is based upon the state-of-the-art generative models. Generative Adversarial Networks (GANs) \cite{goodfellow2014generative} can automatically learn the distribution of the datasets with two adversarial neural networks: a generator network and a discriminator network. The generator network is used to generate the new samples which resemble real data, and the discriminator network tries to discriminate data is real or fake. Conditional Generative Adversarial Networks (cGANs) can deterministically control of output \cite{cgans} by introducing the condition to GANs, such as using the class labels as conditions to generate MNIST images, or using text descriptions as conditions to generate artworks, etc. Our model builds on the "pix2pix" image-to-image framework \cite{pix2pix}, which is based on the cGANS. Figure \ref{fig: structureofnetwork} gives an overview of our calligraphy generation network architecture. Both the generator network and the discriminator network adopt multi-layer deep convolutional networks architectures. The generator network receives a calligraphy image of the standard Kai font as input and generates a corresponding calligraphy image of QiGong font. And a pair of images must be input into the discriminator network in order to determine whether they are from the training set or the generating network.

\subsection{Deep Generative Adversarial Networks}\label{subsec1}
\noindent\textit{\textbf{Generator Network:}} The generator adopts the encoder-decoder architecture as \cite{pix2pix}. The encoder takes an standard Kai font as input and executes downsampling operations through a series of Convolution-BatchNorm-ReLU (CBR) blocks to learn the latent feature representations. Through a succession of Deconvolution-BatchNorm-ReLU (DBR) blocks, the decoder does the opposite operations, reconstructing the QiGong font from the latent feature representations. Each layer of the encoder employs a single skip link to concatenate the layer of the decoder that corresponds to it. The layer $i$ would specifically concatenate all of layer $(n-i)$'s channels, where $n$ is the total number of layers in $G$ network. A lot of low-level information is retained and transmitted through all layers of the $G$ network using the skip connections.

\noindent\textit{\textbf{Discriminator Network:}} The discriminator adopts the structure of \textit{Patch}GAN \cite{pix2pix}. There are numerous CBR blocks in it. The discriminator receives a pair of images as input and outputs a probability indicating whether the images are from the generator or the training set. Each image pair contains a standard Kai font (as condition) and a generated image or ground truth image of QiGong font. 

\subsubsection{Loss Functions}\label{subsubsec1}
The discriminator and generator of our model were trained using a pair of adversarial loss functions $L_d$ and $L_g$. The cGANs original generator loss $L_{gang}$ is combined with $L_1$ loss and $L_{tv}$ loss to create the loss $L_g$: $L_g = L_{gang} + \alpha L_1 + \beta L_{tv}$, where $\alpha$, $\beta$ are weights. Since $L_1$ supports less blurring \cite{pix2pix}, we use $L1$ to measure the difference between the generated image and the ground truth image. To motivate our model to produce more smooth images \cite{crossdomain}, we employ the total variation loss $L_{tv}$. In our paper, the loss functions are expressed as follows:
	
	\begin{equation}
	L_d = \mathbb{E}_{x \in \textbf{s}} log[1 - D(G(x))] - \mathbb{E}_{y \in \textbf{t} } log[D(y)]
	\end{equation}
	
	\begin{equation}
	L_{gang} = -\mathbb{E}_{x \in \textbf{s}} log[D(G(x))]
	\end{equation}
	
	\begin{equation}
	L_{1} = \mathbb{E}_{x \in \textbf{s}, y \in \textbf{t}} [|| y - G(x) ||_1]
	\end{equation}
	
	\begin{equation}
	L_{tv} = \sum_{i, j} ((z_{<i,j+1>} - z_{<i,j>})^2 + (z_{<i+1,j>} - z_{<i,j>})^2)^{\frac{1}{2}}
	\end{equation}
	where \textbf{s} stands for the normal Kai font set and \textbf{t} for the QiGong font set. $z_{<i,j>}$ stands for the $<i_{th}, j_{th}>$ pixel value in the created images, and $z$ stands for the generated image of the QiGong font.

\subsubsection{Evaluation Metrics}\label{subsubsec2}
In this section, we proposed a novel approach to evaluate the generated calligraphy of QiGong font-Coverage Rate (\textit{CR}), which is used to calculate the maximal overlap between the generated images and the ground truth images. Compared to other popular methods for evaluating Chinese calligraphy, such as Mean Absolute Error (MAE), \textit{CR} is more accurate. When creating calligraphy images, MAE is used to calculate the average of the absolute intensity differences between all of the pixels, including the background and valid (black and white) pixels. Due to MAE's excessive use of background pixels in its comparison of two photos, it is unable to quantify the created QiGong font's amount with any degree of accuracy. The overlap of valid pixels between the generated images and ground truth images is calculated with \textit{CR}. This can lessen background pixel influence with evaluation results. The MAE is currently unable to handle calligraphy graphics with moving characters adequately. Due to the MAE's calculation of the pixel difference between two photos' matching positions. In order to achieve the highest CR, we covered the ground truth image with the produced image using a sliding window. As a result, CR can be more precise than MAE when evaluating calligraphy images. The \textit{CR} can be modeled as follows:
    
    \begin{equation}
	CR = \frac{N_{valid} - N_{over} - N_{less}}{N_{valid}}
	\end{equation}
	where $N_{valid}$ is the quantity of black (valid) pixels in the ground truth images. $N_{over}$ represents the number of pixels in the overlap that are black in the created image but white in the ground truth image. While $N_{less}$ denotes the number of pixels in the overlap that are black in the ground truth image but white in the generated image.

\begin{figure}[!htbp]
		\centering
		\includegraphics[width=0.8\textwidth]{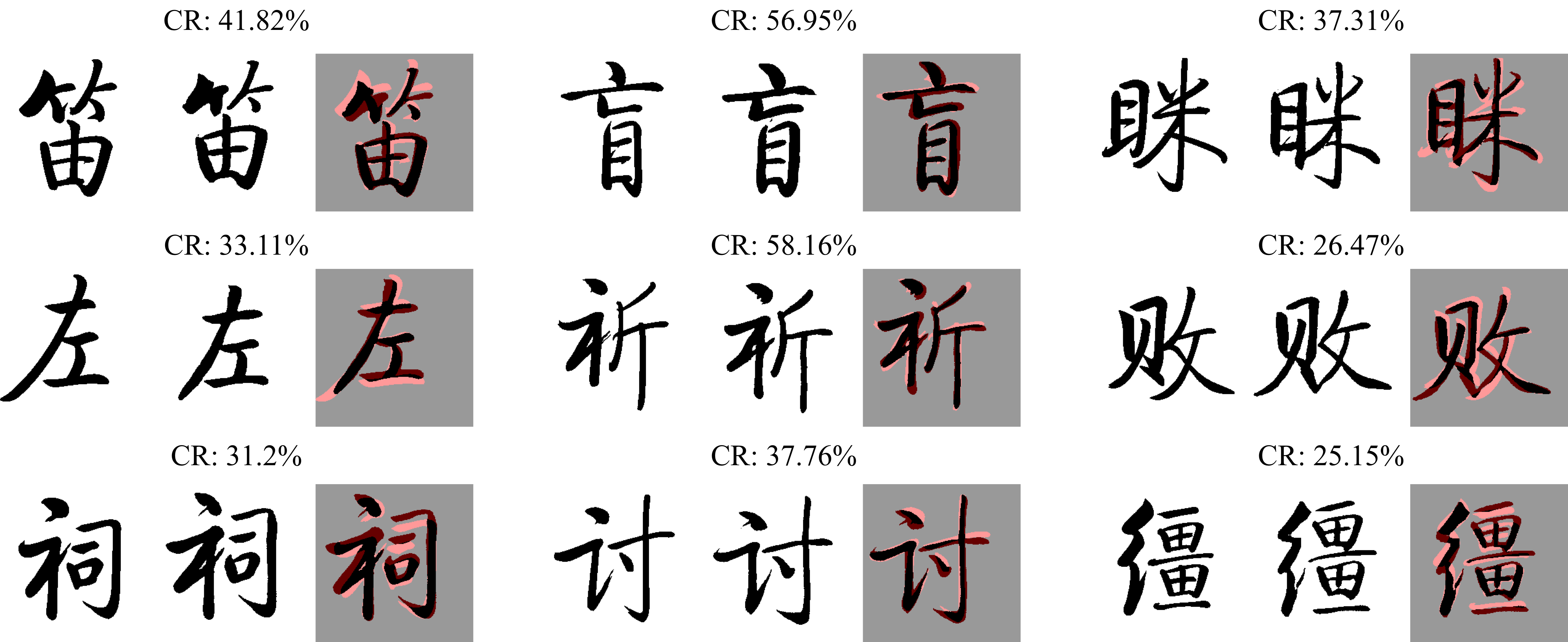}
		\caption{Coverage rate of generated samples.}
		\label{fig: evaluateofcrgoodsamples}
	\end{figure}

\noindent The examples of generated calligraphy images are shown in Figure \ref{fig: evaluateofcrgoodsamples}. Each Chinese character contains three images: the left is the ground truth image, the middle is the generated image, and the right is the overlap image with light grey background. The overlap image is synthesized with the generated image (black) and the ground truth image (red). From these overlap images we can see that the higher the similarity is, the greater the overlap area is. Therefore the CR is relative great. 

	\begin{figure}[!htbp]
		\centering
		\includegraphics[width=0.5\textwidth]{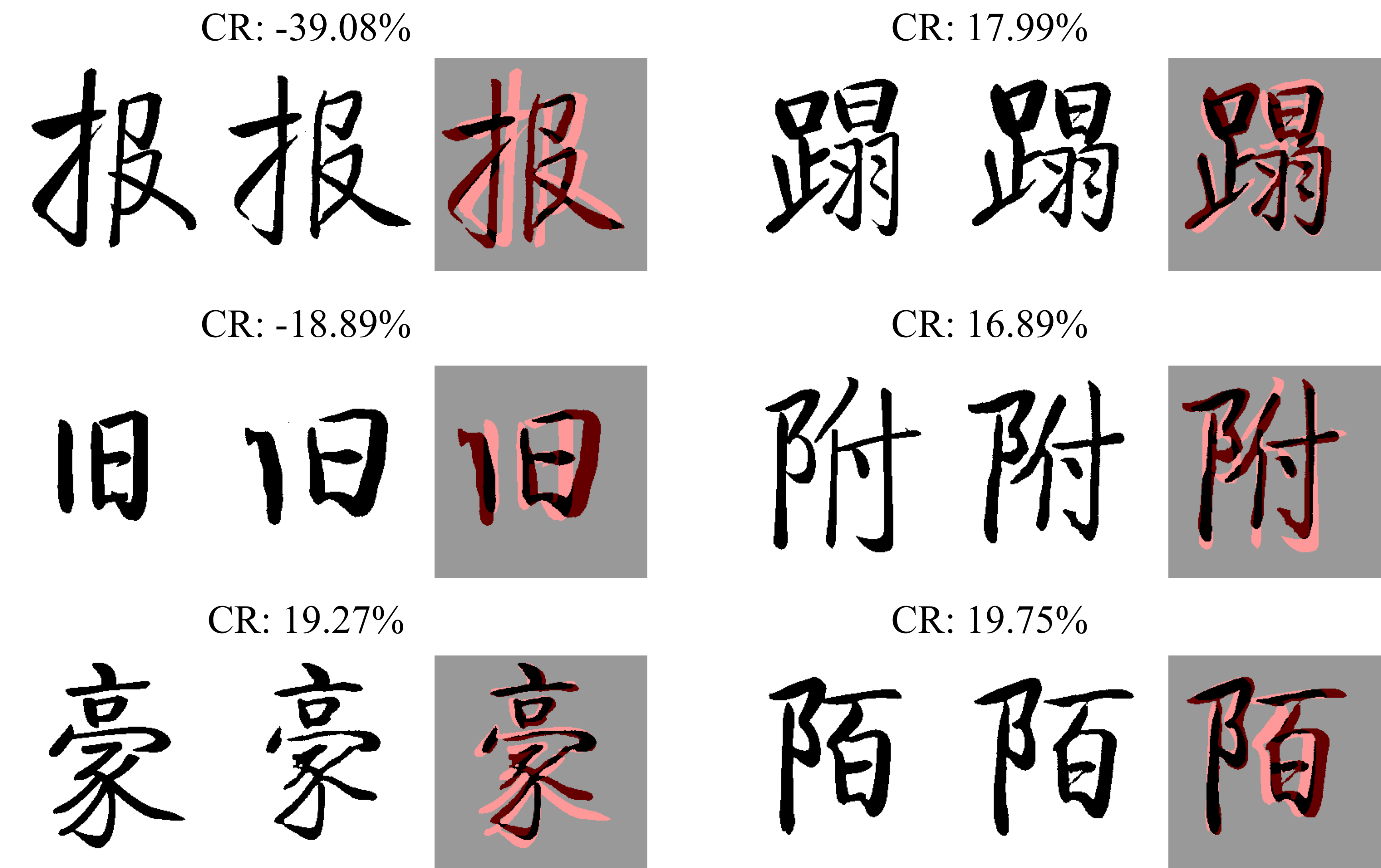}
		\caption{Low coverage rate of samples of generated calligraphy images.}
		\label{fig: crbadsamples}
	\end{figure}

\noindent However, only using the CR to evaluate the calligraphy images is not enough. Specifically, the CR cannot handle the deformation well, including the deformation of overall characters and the local components of characters. Structural deformation of characters can lead to the generated images being unable to cover the ground truth images. The examples of created photos with a low CR are shown in Figure \ref{fig: crbadsamples}. In Figure \ref{fig: crbadsamples}, all characters exist deformation, which lead to reduction of the overlap area. Therefore the CR of those samples are relative low. As a remedy, we introduced another evaluation approach, the Structural Similarity Index Measure(SSIM) \cite{ssim} method, to evaluate how well the created calligraphy image resembles the actual QiGong font image in terms of structure. Image ddegradation is taken into account by the perception-based SSIM model as a perceived change in structural information. SSIM can be expressed as:
	
	\begin{equation}
	SSIM(x, y) = \frac{{(2\mu_x\mu_y + C_1)(2\sigma_{xy}+C_2)}}{{(\mu_x^2 + \mu_y^2+C_1)(\sigma_x^2 + \sigma_y^2 + C_2)}}
	\end{equation}
	where $\sigma_{xy}$ is the covariance of $x$ and $y$, $\mu_x$ and $\mu_y$ are the average values of the pixels in the image $x$ and $y$, respectively, $\sigma_x^2$ is the variance of $x$, and $\sigma_y^2$ is the variance of $y$. Constants $C_1$ and $C_2$ are utilized to prevent a null denominator.

\section{Fonts Dataset and Network Configuration}
\subsection{Dataset Prepartion}
\begin{figure}[!htbp]
		\centering
		\includegraphics[width=0.75\textwidth]{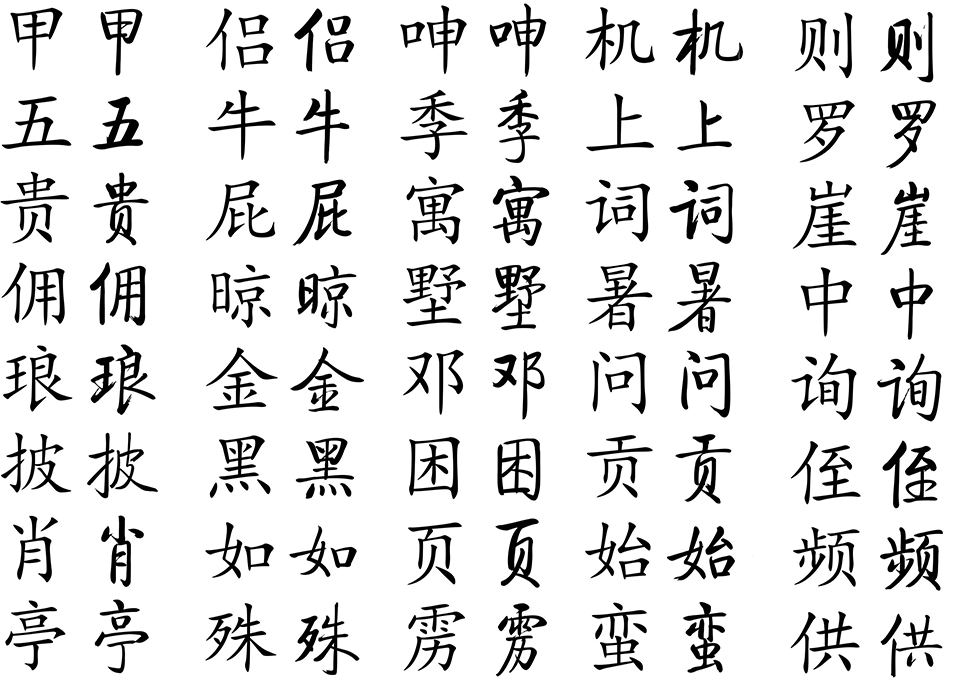}
		\caption{ Samples of dataset. (Left: Kai font. Right: QiGong font.)}
		\label{fig: sampleofdatasets}
	\end{figure}

We will introduce the specifics of dataset preparation in this section. As there are few public datasets of Chinese calligraphic font. Therefore we proposed a new dataset to generate traditional calligraphy images. The proposed dataset contains the Kai font and the QiGong font, which are used as the input font and the target font respectively. in our dataset, each font has at least 3500 images of frequently used characters that have been preprocessed to $256\times256$ binary images. Figure \ref{fig: sampleofdatasets} shows samples from the dataset. Two different fonts from the same character are placed together to construct the image pair as the input of training networks.
	
\noindent So as to generate calligraphy with limited data, the training sets and the testing sets are built which contain at least 500 calligraphy images. To balance the two sets, we choose to manually select the images instead of randomly selecting. The basic requirement of the testing set is that all stroke types of the character appear at least once, as well as all component types. After the above setting, 6 training sets are built by randomly selecting the rest of the images from the dataset, contraining 500, 1000, 1500, 2000, 2500 and 3000 images respectively. And all training sets are set to share the same one testing set.  

\subsection{Network Parameters}
Eight stacked Convolution-BatchNorm-ReLU blocks make up the generator network's encoder, which produces $1\times1$ latent feature representation of the input Kai font calligraphy images. Seven Deconvolution-BatchNorm-ReLU blocks make up the generator network's decoder, which produces $256\times256$ images of calligraphy using the QiGong style font. Each layer of convolution and deconvolution has $5\times5$ filters and a $2\times2$ stride. The discriminator network consists of 4 Convolution-BatchNorm-ReLU blocks with $5\times5$ filters and a $2\times2$ stride, as well as 1 fully-connected layer that outputs the likelihood that the input image pair is real or fake. 

\noindent Before being fed into the network, the training set images are scaled from $256\times256$ to $307\times307$ and then randomly cropped to $256\times256$. We trained the generator and discriminator networks using the Adam optimizer with different momentum terms of Adam \cite{adam}:
$beta_g$ (generator) and $beta_d$ (discriminator). To improve the generator ($lr_g$) and discriminator ($lr_d$), two learning rates are utilized, starting at 0.001. We employ the 0.9 momentum exponential decay of learning rates. Having a 0 mean and 0 standard deviation, the Gaussian distribution is used to initialize all layer parameters. We train our model for 100 epochs due to the limited computing power.

\begin{figure}[htbp]
		\centering
		\includegraphics[width=0.35\textwidth]{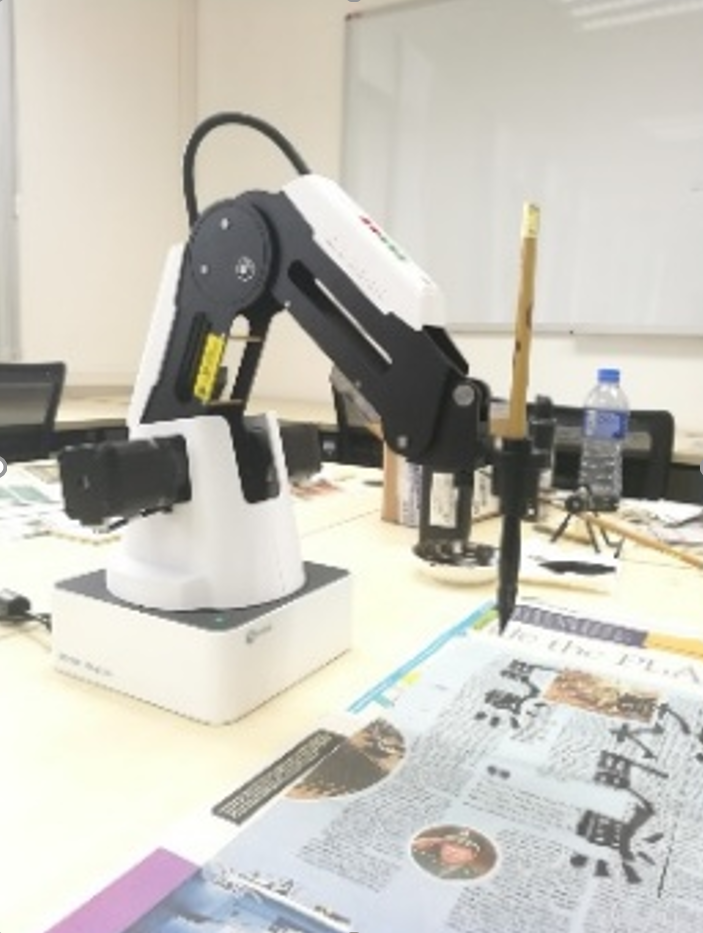}
		\caption{The calligraphy robot-DOBOT Magician.}
		\label{fig: dobot}
    \end{figure}

\noindent The experiment was set up on a PC with an Intel i7 processor running at 2.8 GHz, 16.0 GB of RAM, and the Linux Server 14.04 operating system. TensorFlow was used to create our model With CUDA 8.0 and cuDNN 5.1. We execute our model on an Nvidia GeForce GTX 1070-8GB GPU. Our model could be trained using the training set of 3000 and 100 epochs in about 20 hours. The PC is equipped with a calligraphy robot DOBOT Magician \ref{fig: dobot} for subsequent robotic writing.

\section{Experiment Results}
This section assesses and discusses the output quality of the QiGong font. We train our model on different training sets, and evaluate the quality of generated images, including CR, SSIM and visual effect. At last, we construct a Turing test contains 50 generated images and 50 ground-truth images. 
	
\subsection{Overview of the Final Results}
\begin{figure}[!htbp]
		\centering
		\includegraphics[width=1\textwidth]{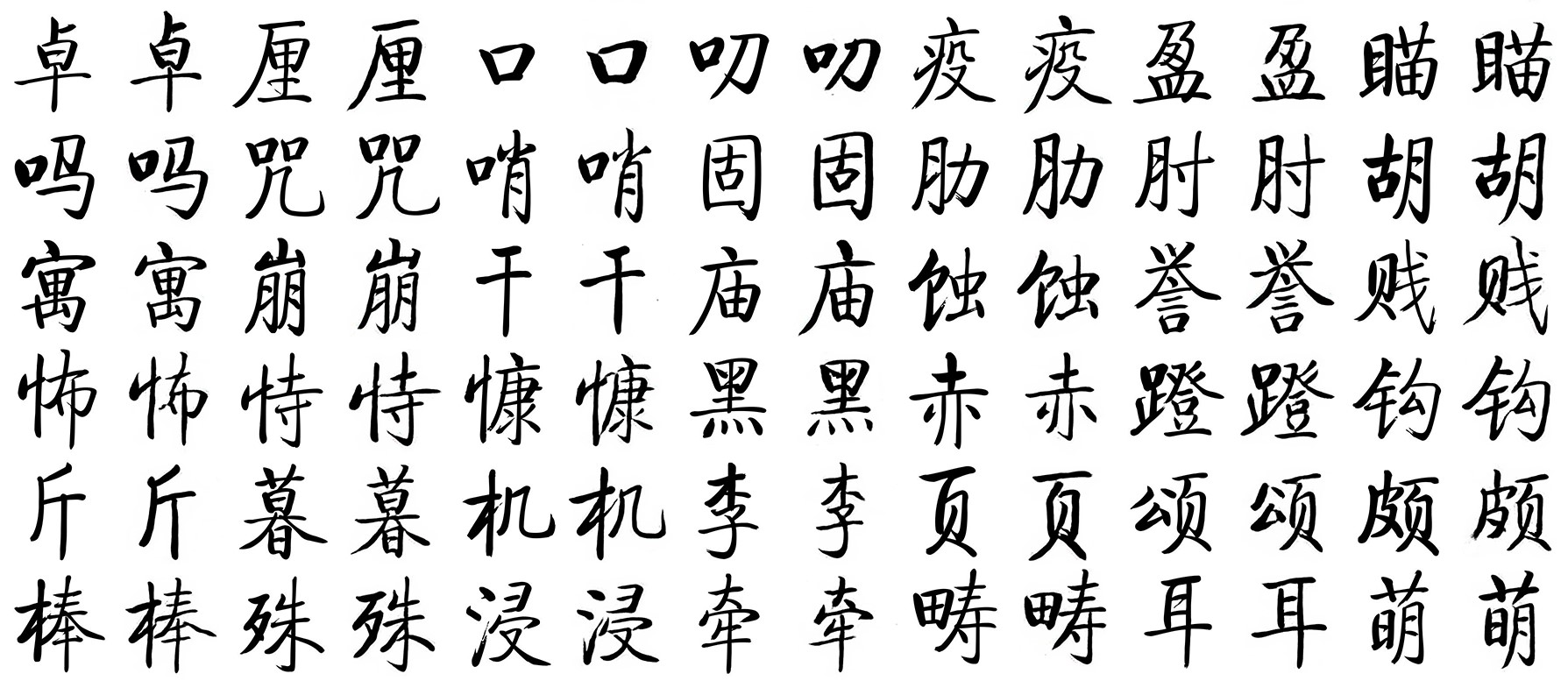}
		\caption{Overview of generated results. Each image pair of characters contains the ground-truth image (Left) and the generated image (Right).}
		\label{fig: summaryofresultslowres}
	\end{figure}

The overview of the calligraphy images that our model produced is shown in figure \ref{fig: summaryofresultslowres}. The results can meet the basic requirements of Chinese calligraphy, including correct strokes of each Chinese character, suitable size, good and smooth thickness of strokes and the satisfactory beginning and ending details of strokes, etc. Even the complex characters and complex component structure of characters, our model can handle well. The calligraphy images of our model generated are comparable to those people who have certain knowledge of Chinese calligraphy. Besides, our generated calligraphy images of QiGong font are similar to the ground truth images of QiGong font at a glance, including global shape, size and local strokes of characters. Some strokes of characters in the generated images are very similar the ground truth images. 

\subsection{Output Threshold} 
    All images in training sets are pre-processed to $256 \times 256$ binary images before feeding into our model (0 represents the white background pixel and 1 is the black valid pixel). The outputs of our model are the $256 \times 256$ matrices of probabilities that the pixel is black. In order to determine whether the pixel is black, we used the threshold to filter the generated calligraphy images of our model. 

\begin{figure}[!htbp]
		\centering
		\includegraphics[width=0.75\textwidth]{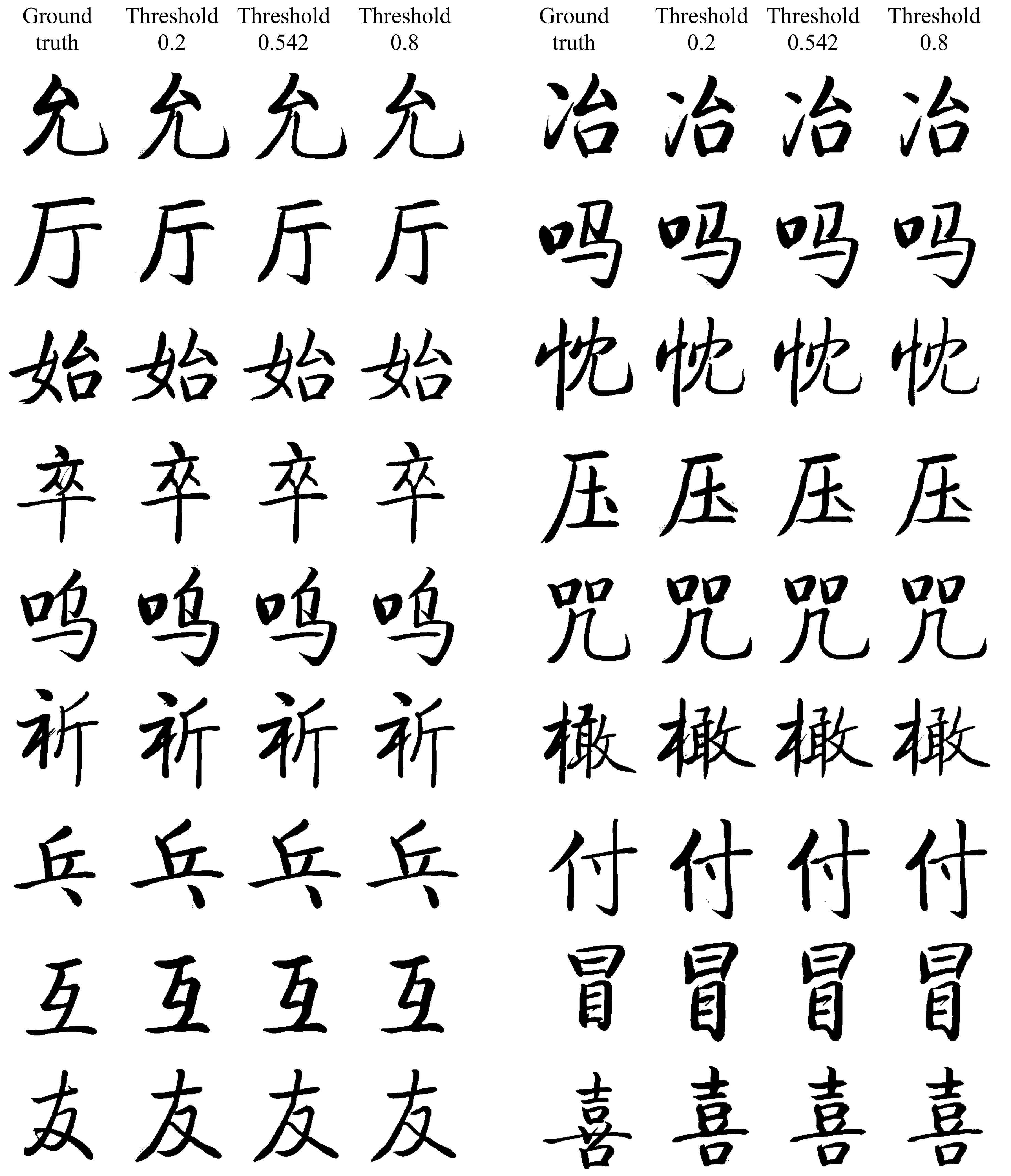}
		\caption{Generated samples under the different threshold setting.}
		\label{fig: threshold0542}
	\end{figure}

\noindent We choose the threshold based on the number of valid pixels of the ground truth images. Because there should be as many valid pixels in the generated image as there are in the ground truth image, each generated image has an independent threshold that the number of probability larger than this threshold is close to the number of valid pixels of ground-truth image. We use the average of independent thresholds of whole generated images as the global threshold. In this paper, the global threshold of all generated images of QiGong font is 0.542. Figure \ref{fig: threshold0542} shows samples of generated images in these three groups with different thresholds.

    \subsection{Results on the different sizes of the training sets} 
\begin{figure}[htbp]
        \centering
        \subfloat[][CR]{\includegraphics[width=0.5\textwidth]{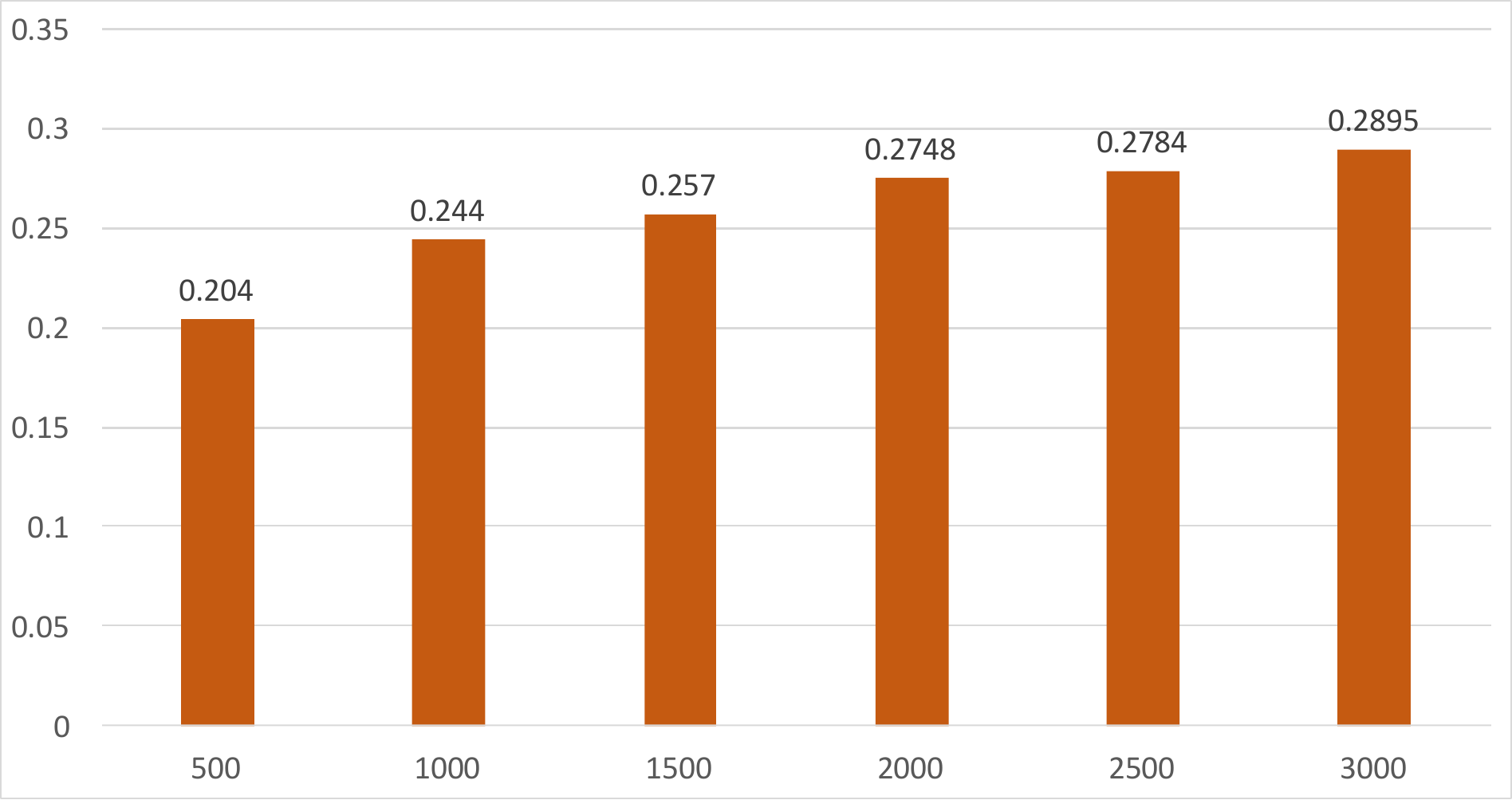} \label{fig: avgcroftraningsets}}
        \subfloat[][SSIM]{\includegraphics[width=0.5\textwidth]{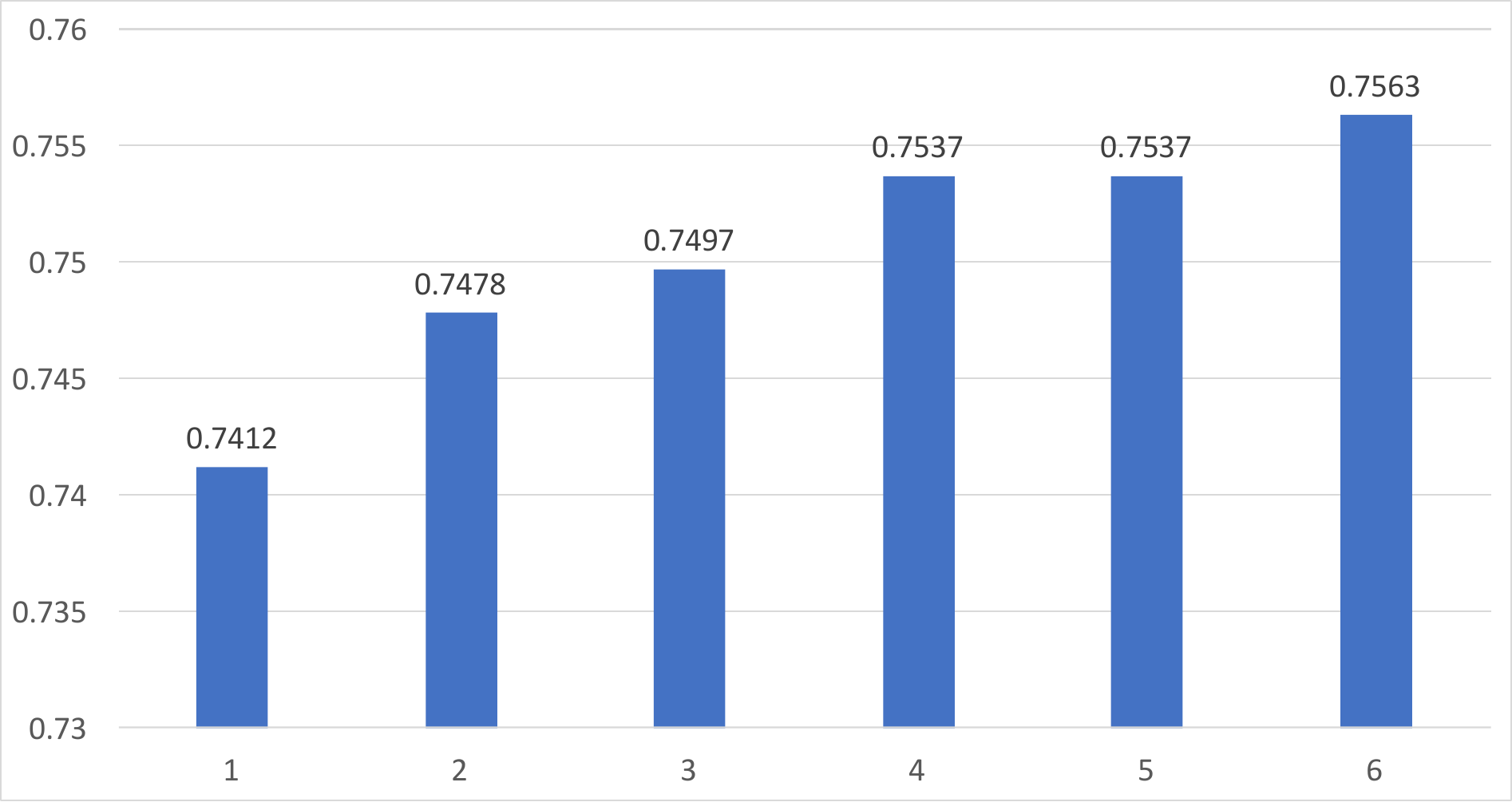} \label{fig: avgssimoftrainingsets}}
        \caption{CR and SSIM of generated calligraphy on different training sets.}
        \label{fig: avgcrssimoftrainingsets}
    \end{figure}

    In order to determine how many Chinese characters can essentially satisfy the requirements of QiGong font generation, we used various size training sets. The average CR and SSIM of the test set's images are shown in the figure \ref{fig: avgcrssimoftrainingsets}. The CR and SSIM of generated images have improved slightly with an increase in training samples. Figure \ref{fig: avgcrssimoftrainingsets} shows that the increases in CR and SSIM occur when there are 500 to 2000 training sets. When between 2500 and 3000 training data were used in the training, the increases, however, tapered down. As a result, training set 2000 essentially satisfies the criteria for the creation of the QiGong font.
    
\begin{figure}[!htbp]
	    \centering
	    \includegraphics[width=0.5\textwidth]{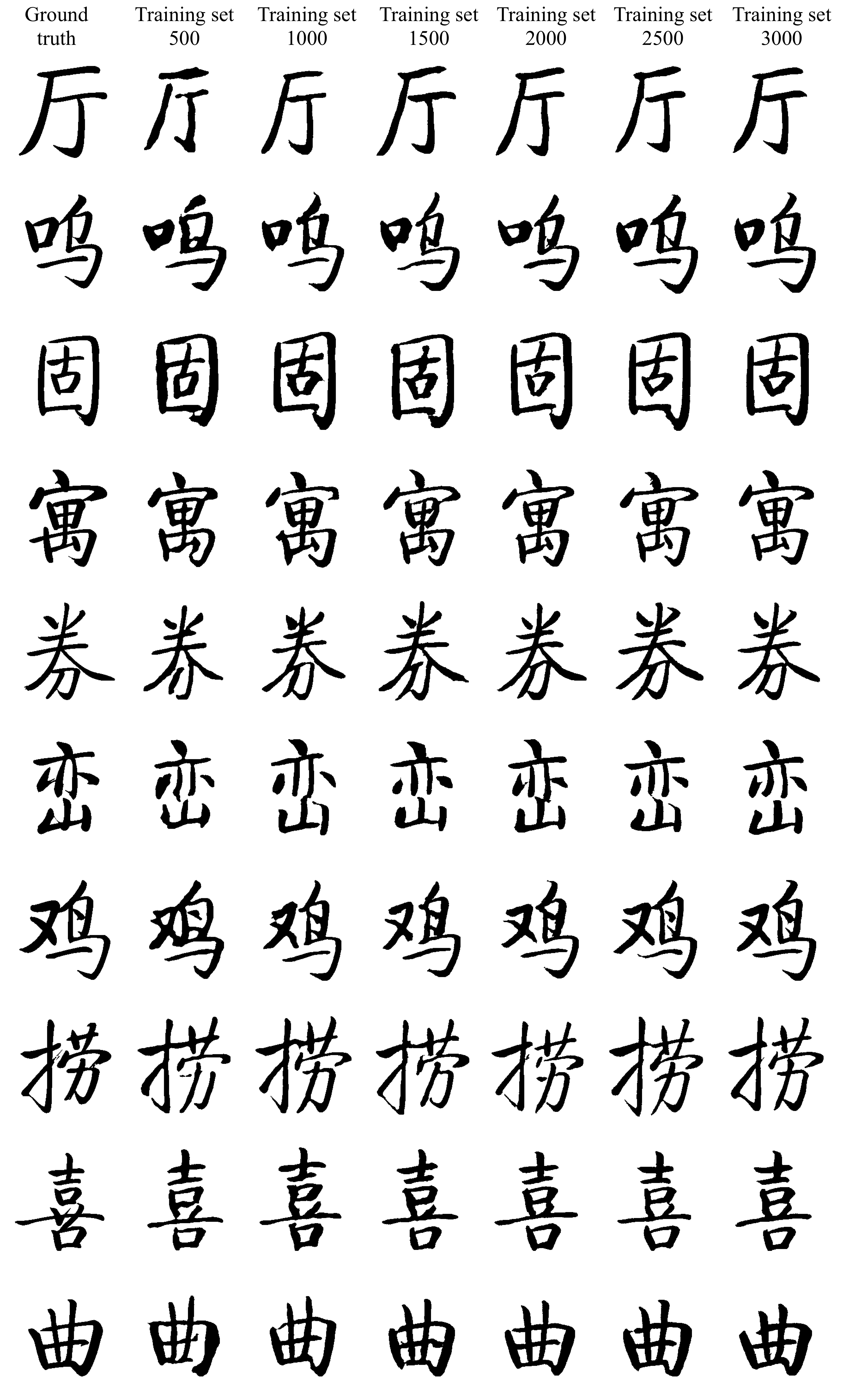}
	    \caption{CR of generated calligraphy on different training sets.}
	    \label{fig: samplesoftrainingsets}
	\end{figure}

\noindent Figure \ref{fig: samplesoftrainingsets} displays examples of the QiGong font's created calligraphy images, which range in size from 500 to 3000. The actual copies of the QiGong font are in the first column. Figure \ref{fig: samplesoftrainingsets} shows that as the number of training samples increases, the quality of the generated images gradually rises. The training set 500 generated photographs have the worst quality. Chinese letters essentially only exist in their shape; their fine structures are unclear. Character strokes have missing portions. The strokes themselves aren't smooth enough, though. The generated images from training set 3000 have the best quality, including excellent shape and fluid character strokes. The generated images on training set 2000 essentially match the requirements of the QiGong font creation in this paper, and the visual variations between the generated images on training sets 2000, 2500, and 3000 are negligible.

\begin{figure}[htbp]
        \centering
        \subfloat[Training set 500]{\includegraphics[width=0.5\textwidth]{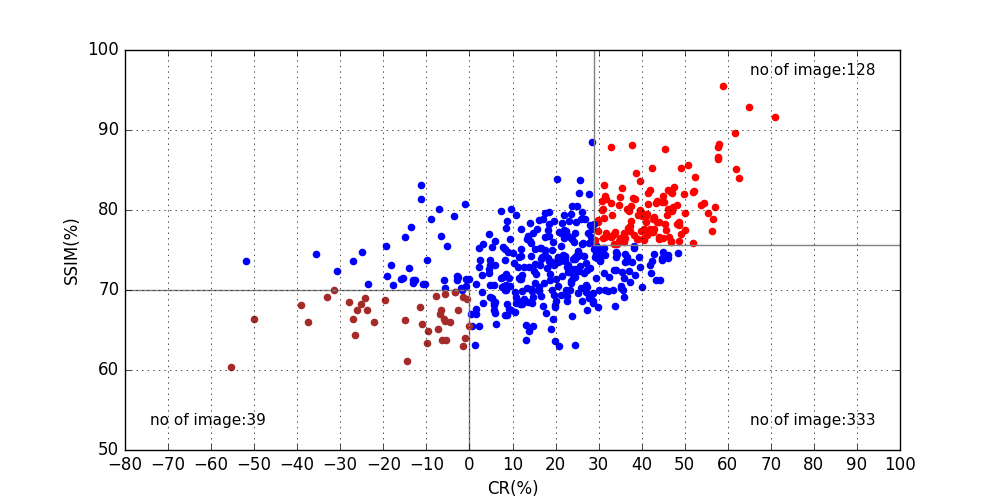} \label{fig: qualityscatterplotsof500}}
        \subfloat[Training set 1000]{\includegraphics[width=0.5\textwidth]{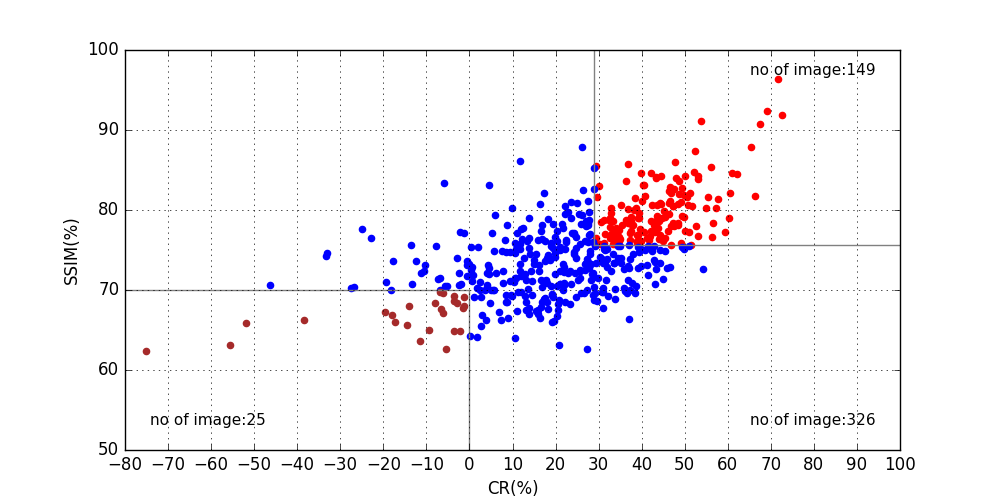} \label{fig: qualityscatterplotsof1000}}\hfill
        
        \subfloat[Training set 1500]{\includegraphics[width=0.5\textwidth]{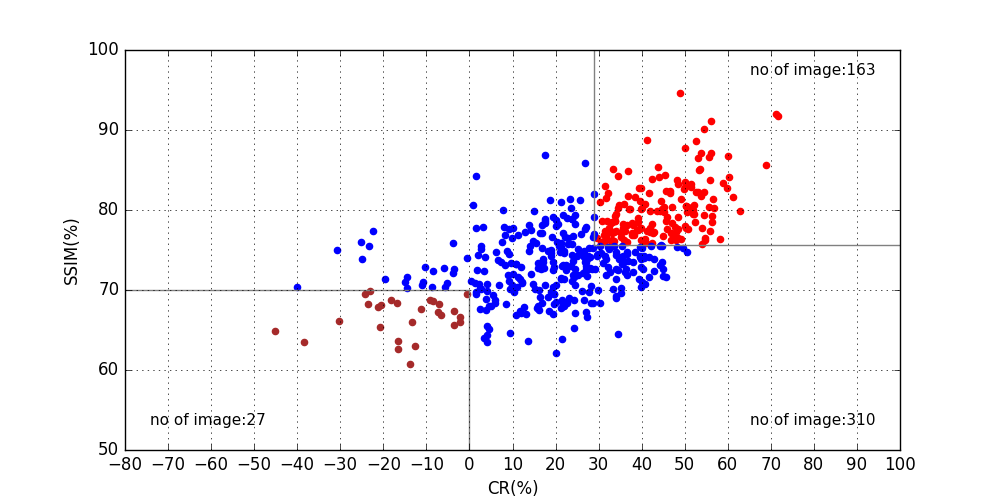} \label{fig: qualityscatterplotsof1500}}
        \subfloat[Training set 2000]{\includegraphics[width=0.5\textwidth]{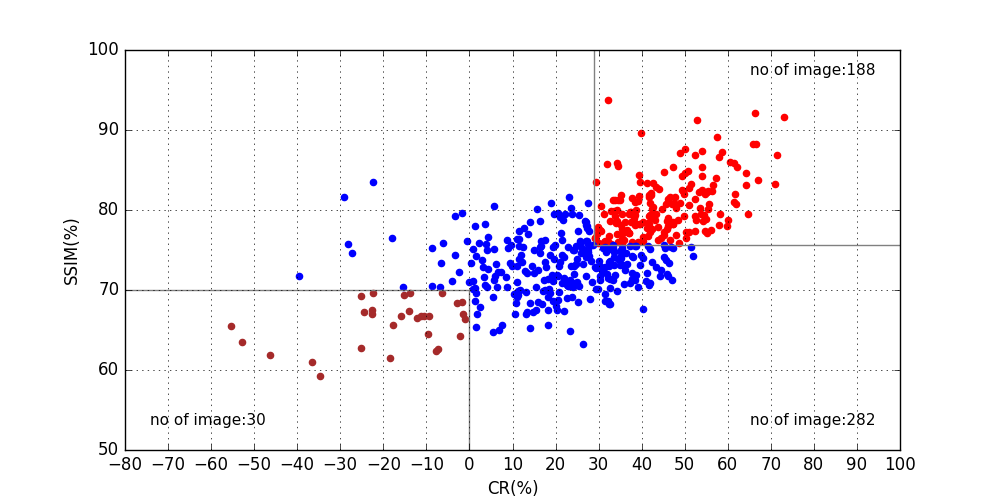} \label{fig: qualityscatterplotsof2000}} \hfill
        
        \subfloat[Training set 2500]{\includegraphics[width=0.5\textwidth]{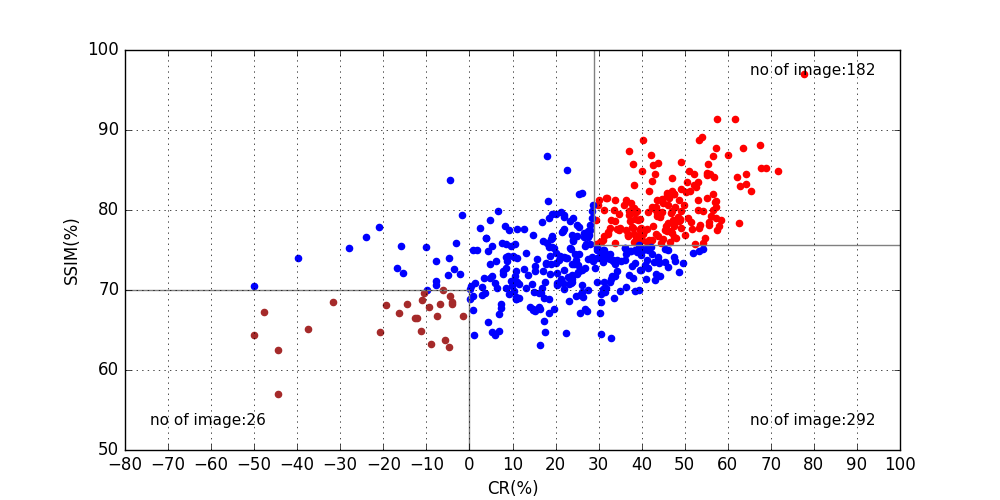} \label{fig: qualityscatterplotsof2500}}
        \subfloat[Training set 3000]{\includegraphics[width=0.5\textwidth]{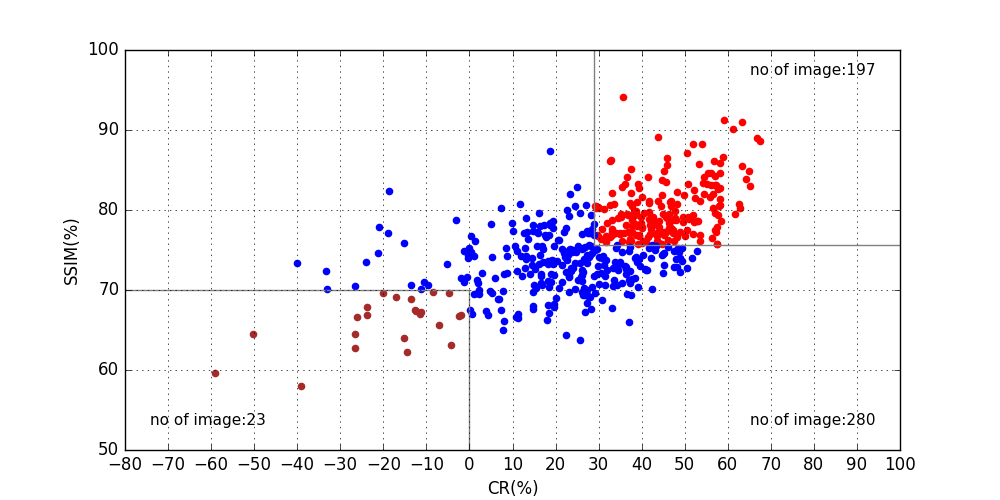} \label{fig: qualityscatterplotsof3000}}
         
        \caption{The quality division of generated calligraphy on different training sets.}
        \label{fig: qualityscatterplots}
    \end{figure}

\noindent {\textbf{Samples of different quality of generation images.}} 
With using the CR and SSIM, we quantify the QiGong font generating images. Two criteria are used in this study to evaluate the caliber of the QiGong font generation images. On the training set of 3000 images, the average CR and SSIM is one. In this research, we found that the average SSIM is $75.91\%$ and the average CR is $30.25\%$. Another is that SSIM is $68.0\%$ and CR is $0.0 \%$. High quality generation images are those with $CR > 30.25\%$ and $SSIM > 75.91\%$, whereas low quality images are those with $CR < 0.0\%$ and $SSIM < 68.0\%$. The images from the remaining generations are of medium quality. Figure \ref{fig: qualityscatterplots} displays the outcomes of dividing calligraphy images of varying quality generated from several training sets. The middle and low quality images of the generation are proportionally reduced as the training sample size rises while the high quality images increase.

\begin{figure}[htbp]
        \centering
        \subfloat[][High]{\includegraphics[width=0.33\textwidth]{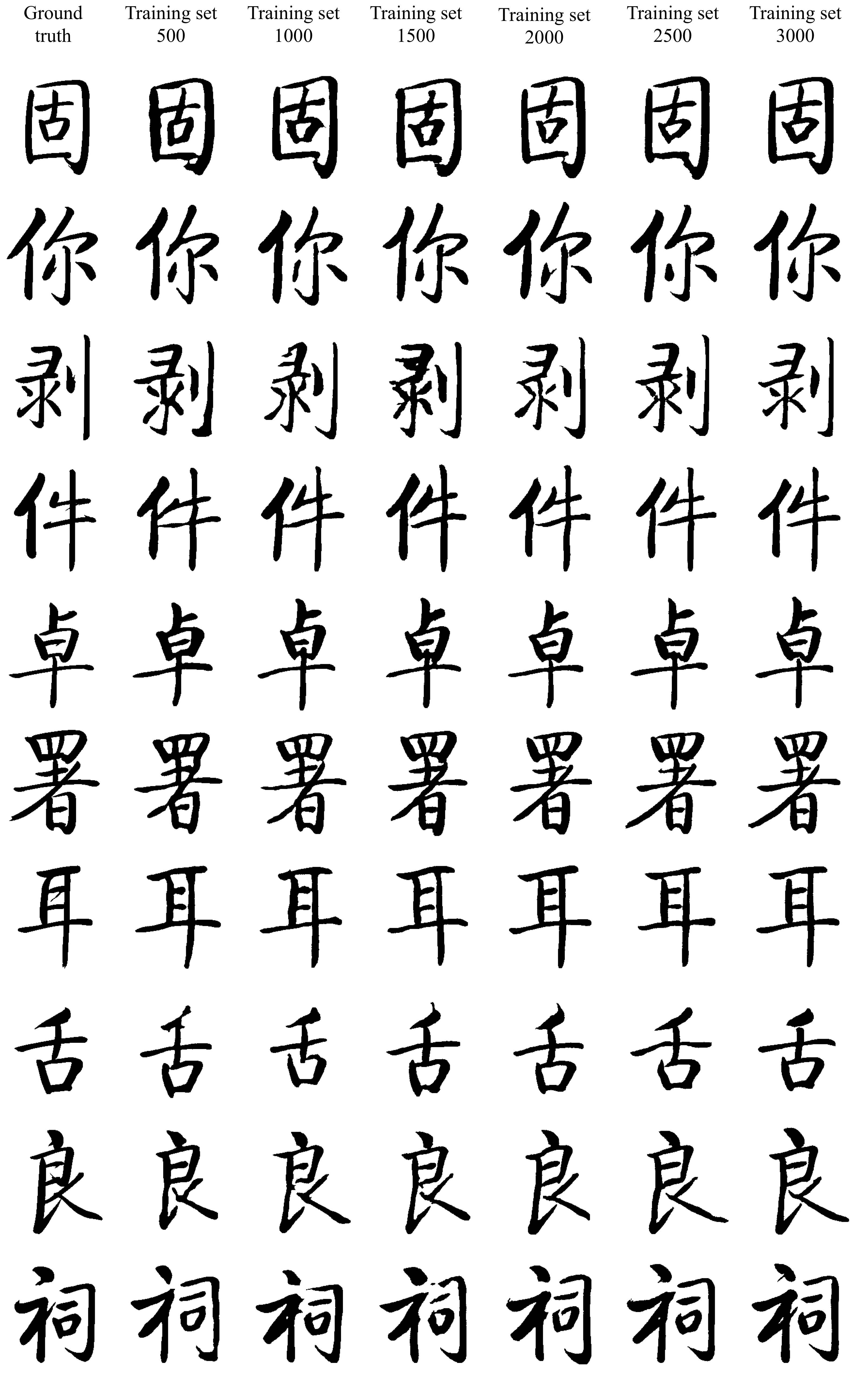} \label{fig: goodsamplesbysize}}
        \subfloat[][Medium]{\includegraphics[width=0.33\textwidth]{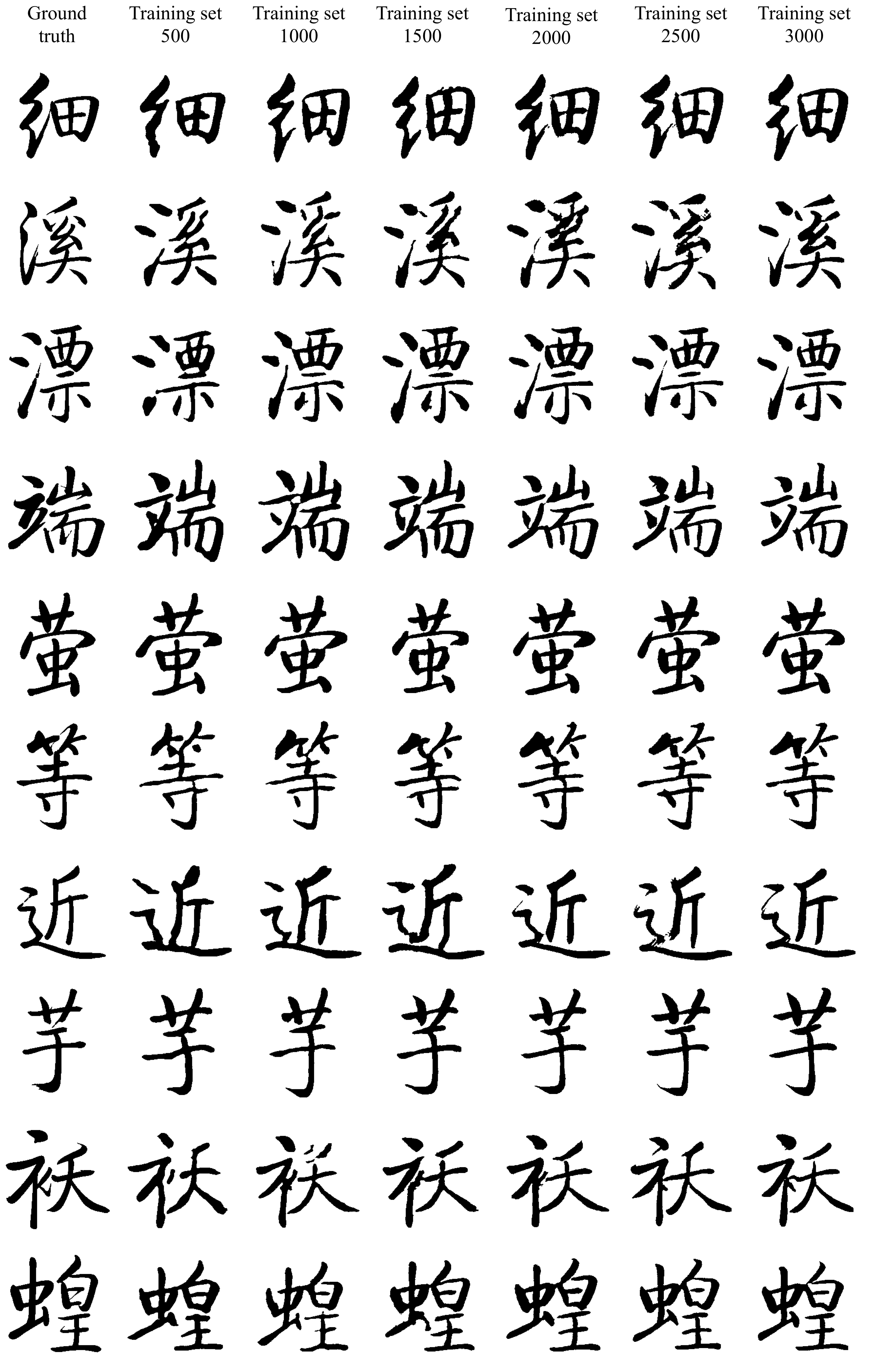} \label{fig: middlesamplesbysize}}
        \subfloat[][Low]{\includegraphics[width=0.33\textwidth]{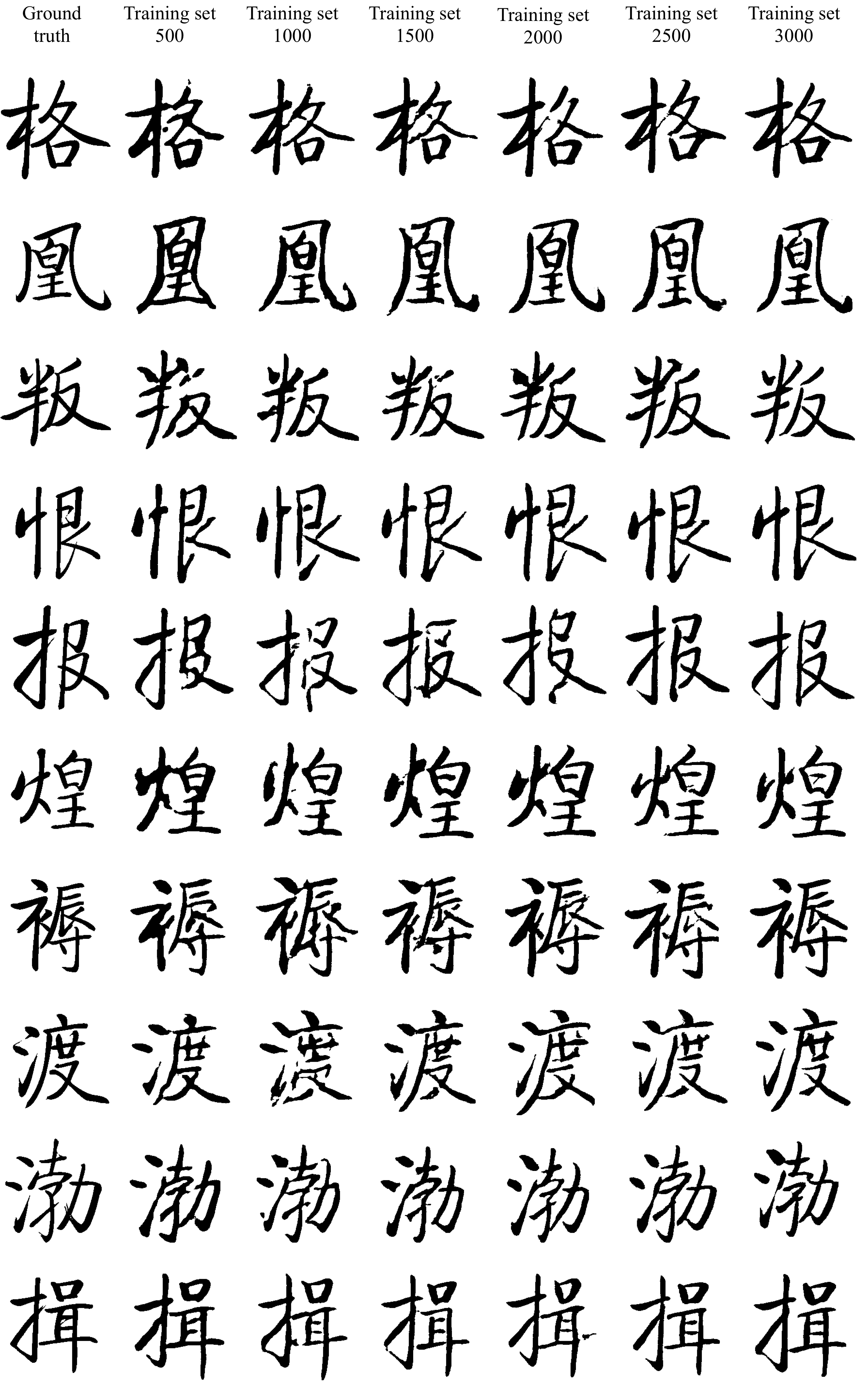} \label{fig: badsamplesbysize}}
        \caption{High, medium and low quality of generated calligraphy samples.}
        \label{fig: various quality of samples}
    \end{figure}

\noindent Figure \ref{fig: various quality of samples} show the high, middle and low quality of generation images of QiGong font respectively. As the training samples increased, the quality of all generation images become better and better. 

	
	
    \subsection{Turing Test} 
\begin{figure}[!htbp]
	    \centering
	    \includegraphics[width=0.7\textwidth]{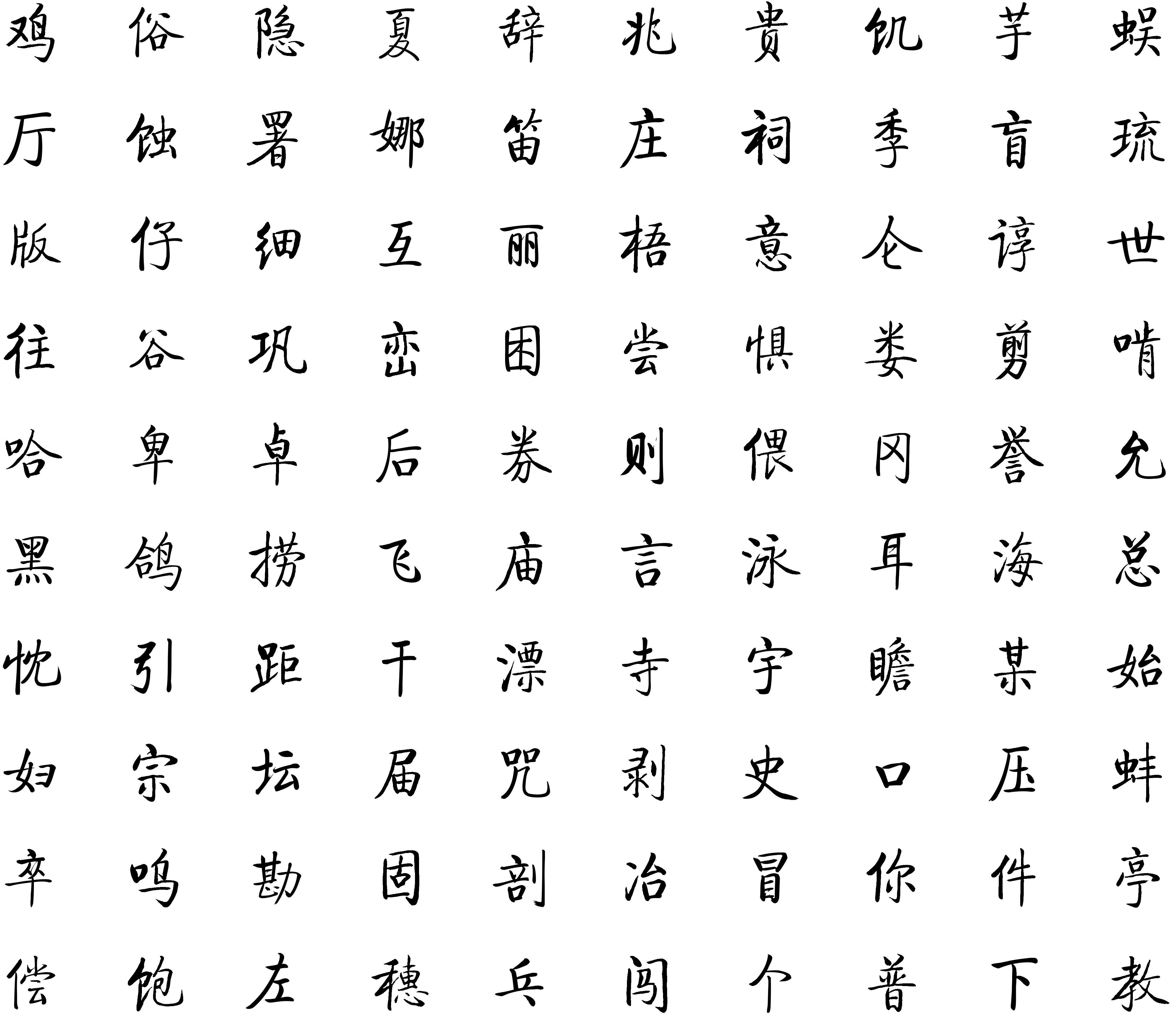}
	    \caption{The Turing test template.}
	    \label{fig: turingtest}
	\end{figure}

\begin{figure}[!htbp]
	    \centering
	    \includegraphics[width=0.7\textwidth]{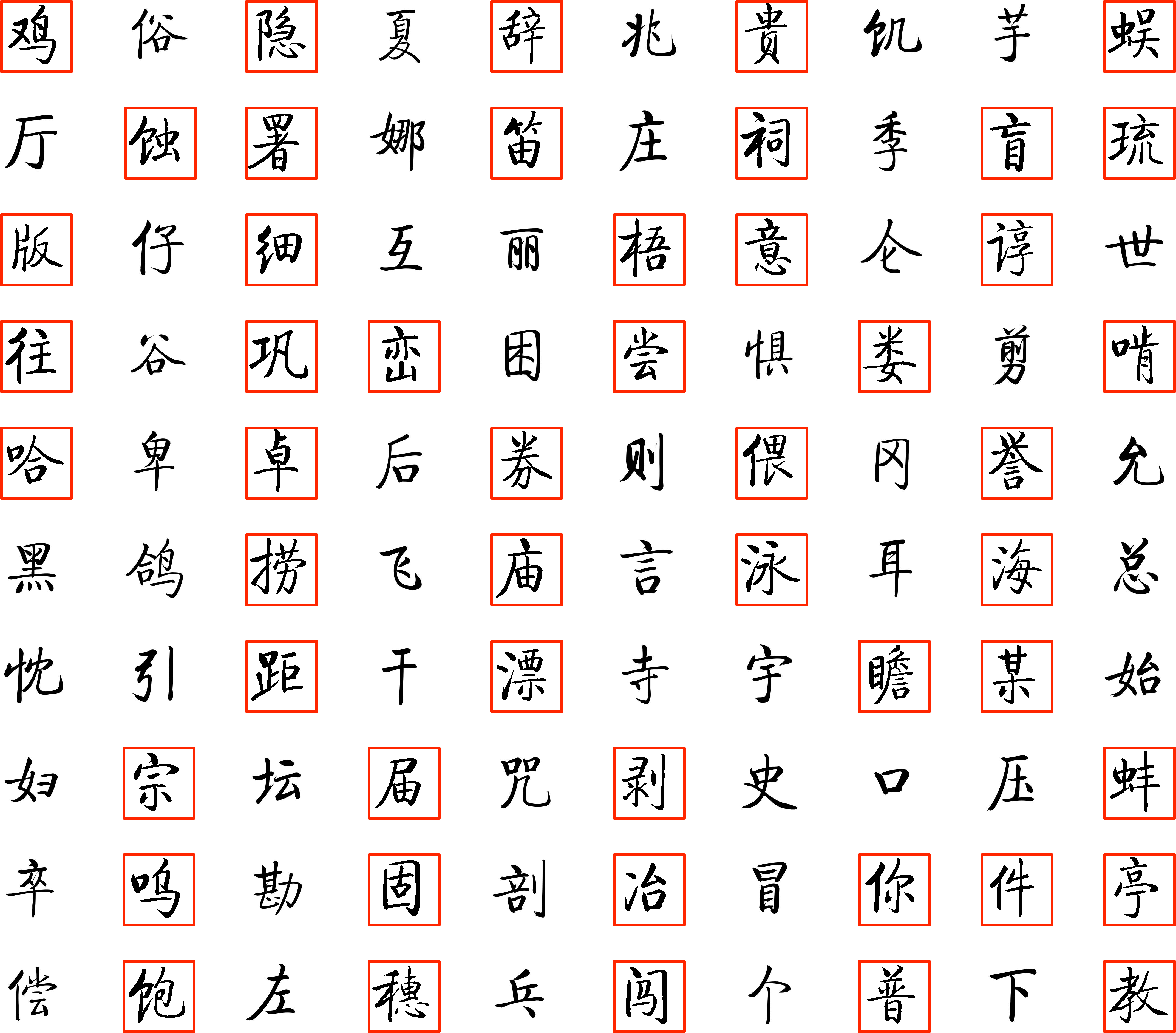}
	    \caption{The answer example of Turing test.}
	    \label{fig: answerofturingtest}
	\end{figure}

    To quantitatively measure the similarity of generation images of QiGong font and the ground truth images, Turing test is conducted (Figure \ref{fig: turingtest}). 100 images of QiGong font are randomly chosen and mixed in one testing paper, including 50 generation images and 50 ground truth images. Figure \ref{fig: answerofturingtest} shows the answer of testing, in which all the generation images of QiGong font are circled by red squares. 
	
\noindent We invited 30 people with a certain knowledge of Chinese calligraphy to participate in the Turing test. Each participant is asked to pick out the generation images. The average accuracy of distinguishing the generation images from the ground truth images is 56\%. The result of Turing test indicates that the generation images of QiGong font, for both the overall style and the details of strokes, are hard to be distinguished from the ground truth images.
    
	
	
	\subsection{Font Style Transfer from Kai to SimSun}
\begin{figure}[!htbp]
        \centering
        \subfloat[l][Training set 500]{\includegraphics[width=0.4\textwidth]{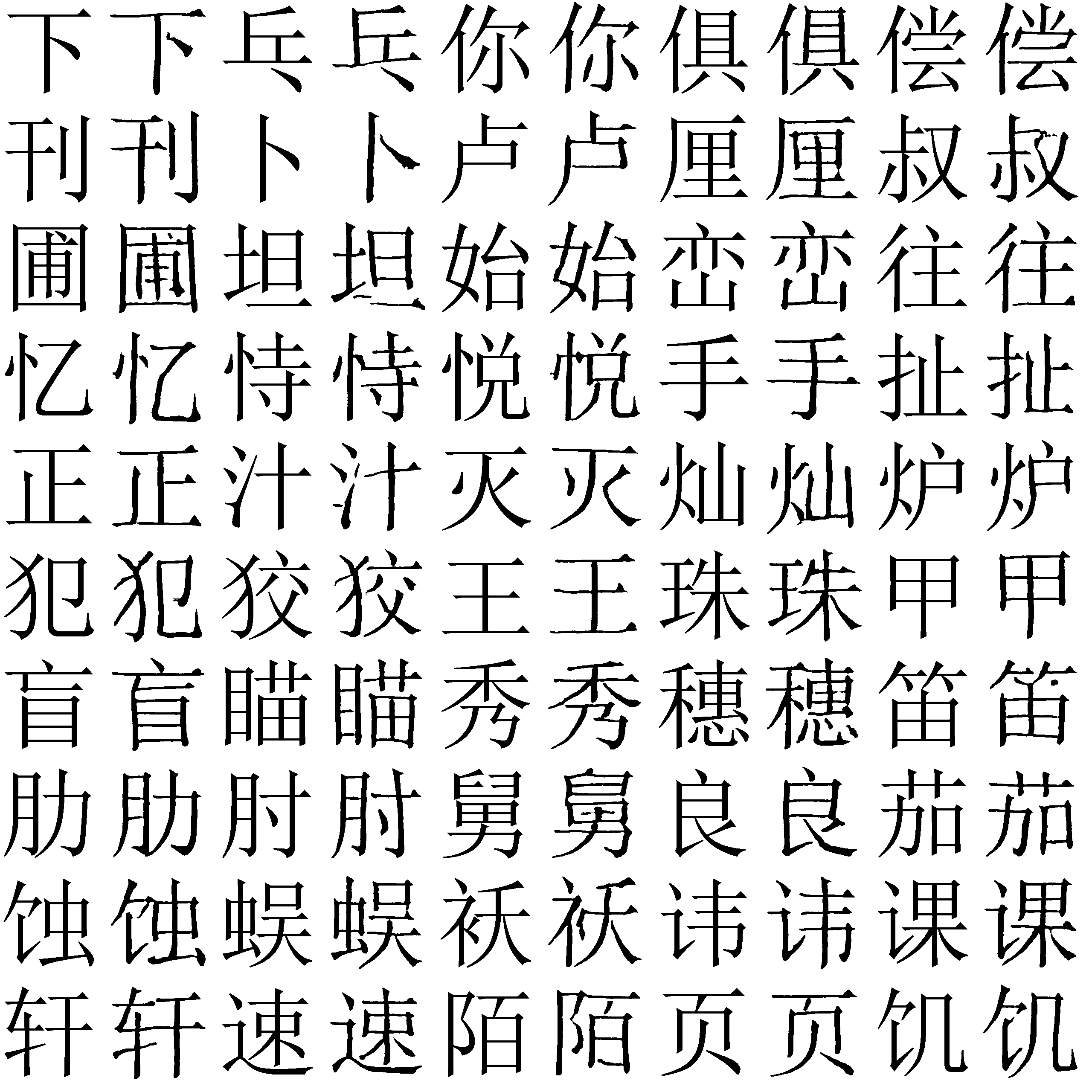} \label{fig: simsun500}}
        \subfloat[r][Training set 1000]{\includegraphics[width=0.4\textwidth]{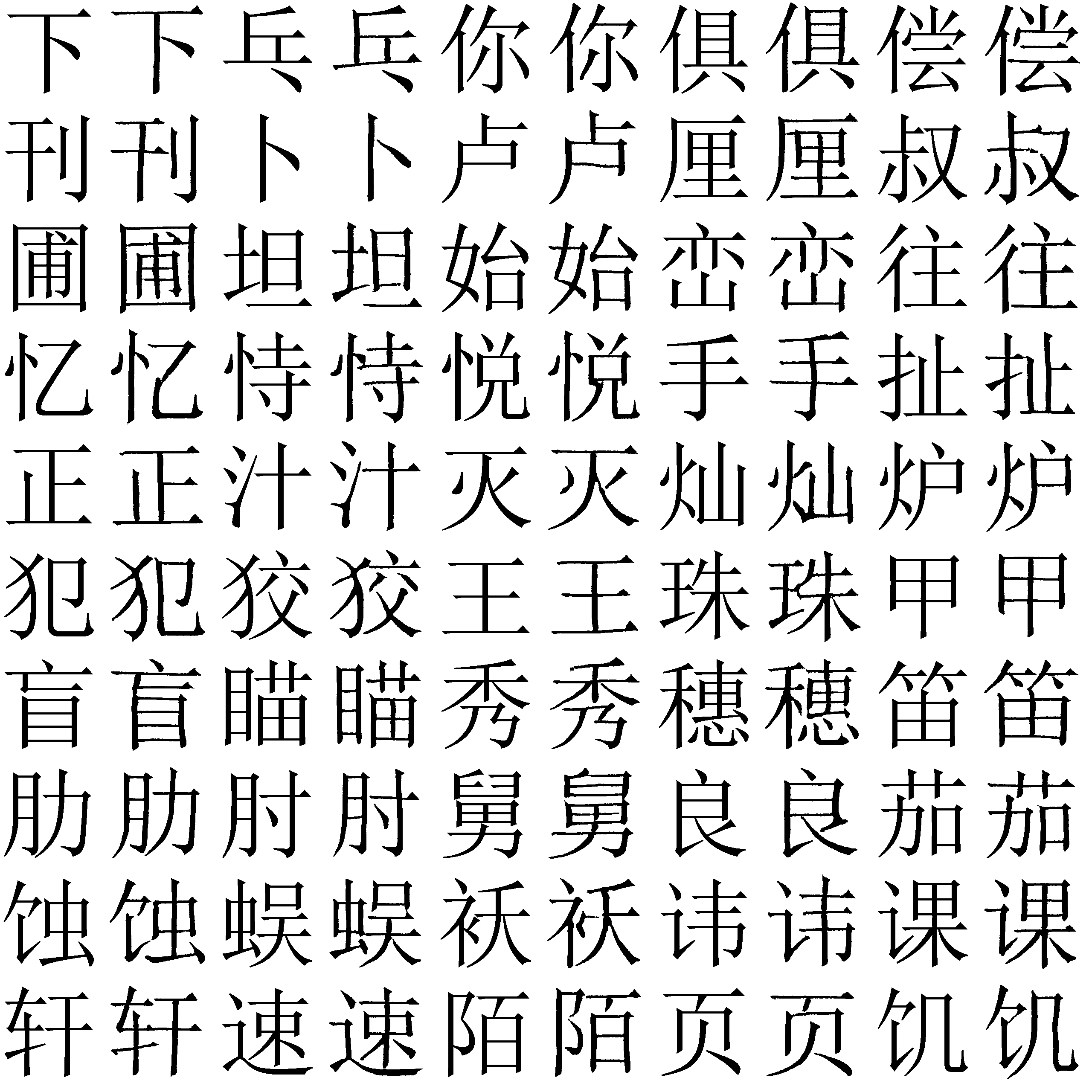} \label{fig: simsun1000}}\hfill
        
        \subfloat[l][Training set 1500]{\includegraphics[width=0.4\textwidth]{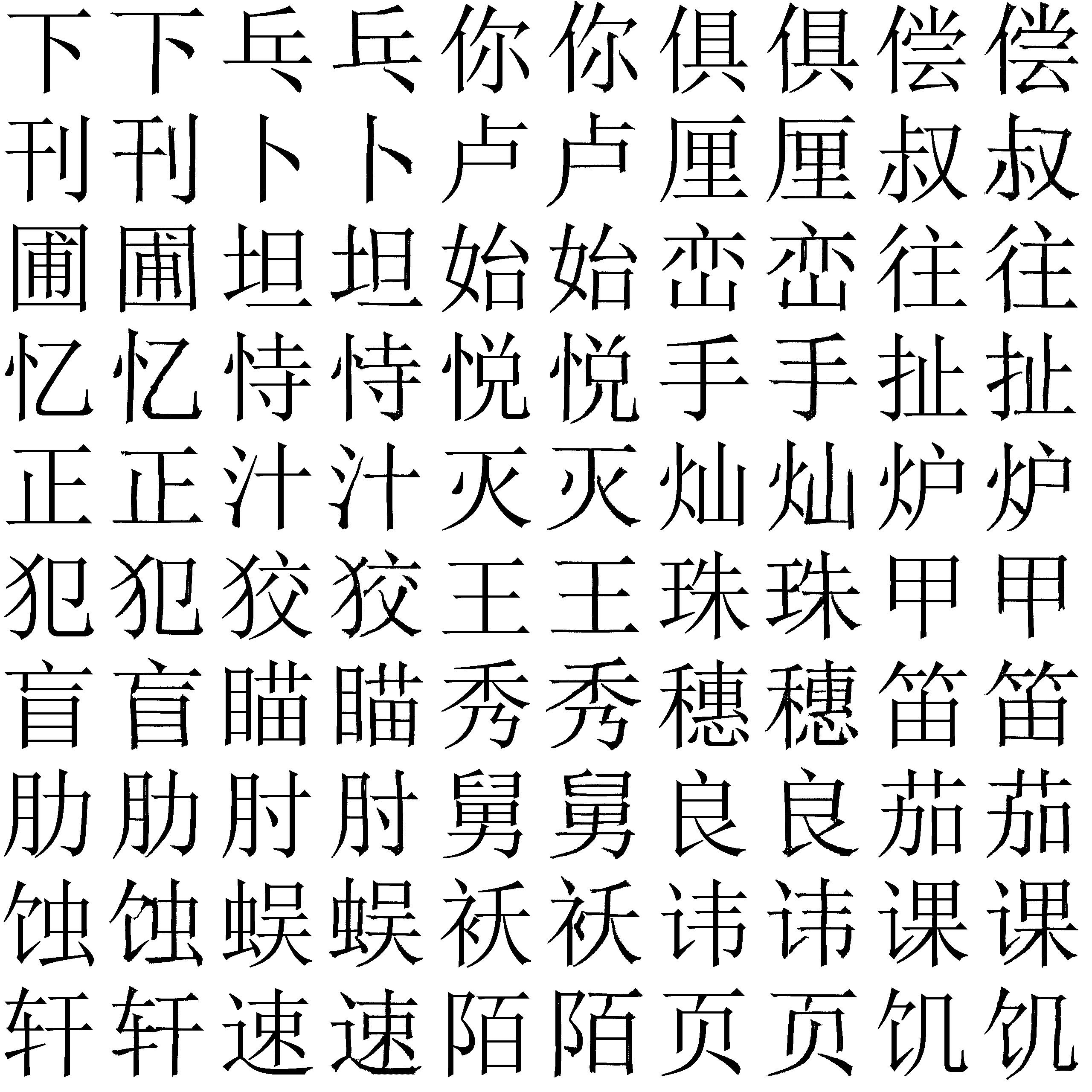} \label{fig: simsun1500}}
        \subfloat[r][Training set 2000]{\includegraphics[width=0.4\textwidth]{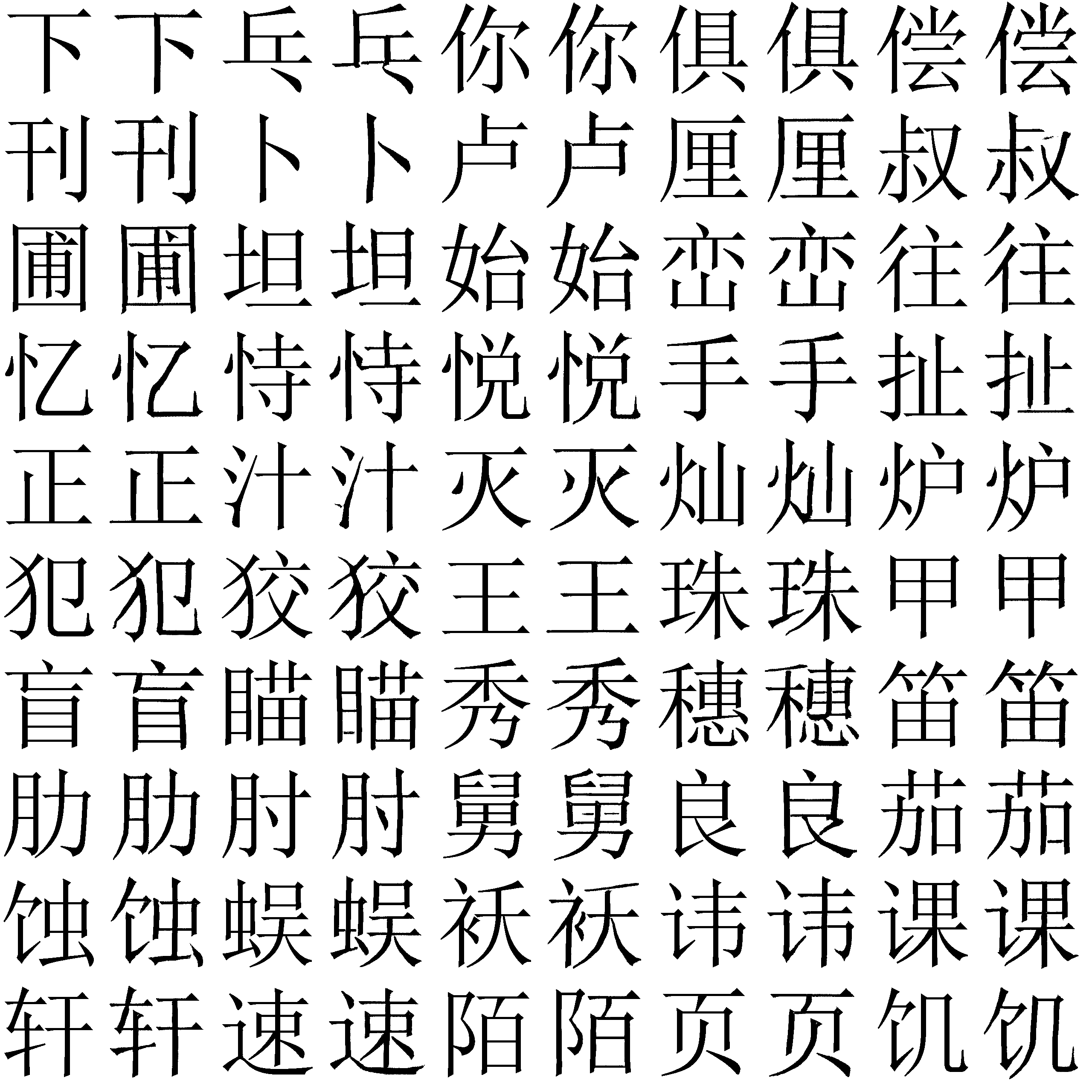} \label{fig: simsun2000}} 
        
        \caption{Generated results of SimSun font on different training sets. Each contains 50 image pairs, in which the left is the ground truth image while the right is the generated image.}
        \label{fig: simsunfont}
    \end{figure}

	In this paper, we used our approach to generate the calligraphy images of SimSum font with the standard Kai font as input. We adopted 4 training sets containing 500, 1000, 1500 and 2000 images of SimSun font respectively. Figure \ref{fig: simsunfont} shows the generation images of SimSun font on different training sets. Our approach also achieves desirable results of generating the SimSun font.
	
	\subsection{Limitations} 
\begin{figure}[!htbp]
        \centering
        \subfloat[l][]{\includegraphics[width=0.3\textwidth]{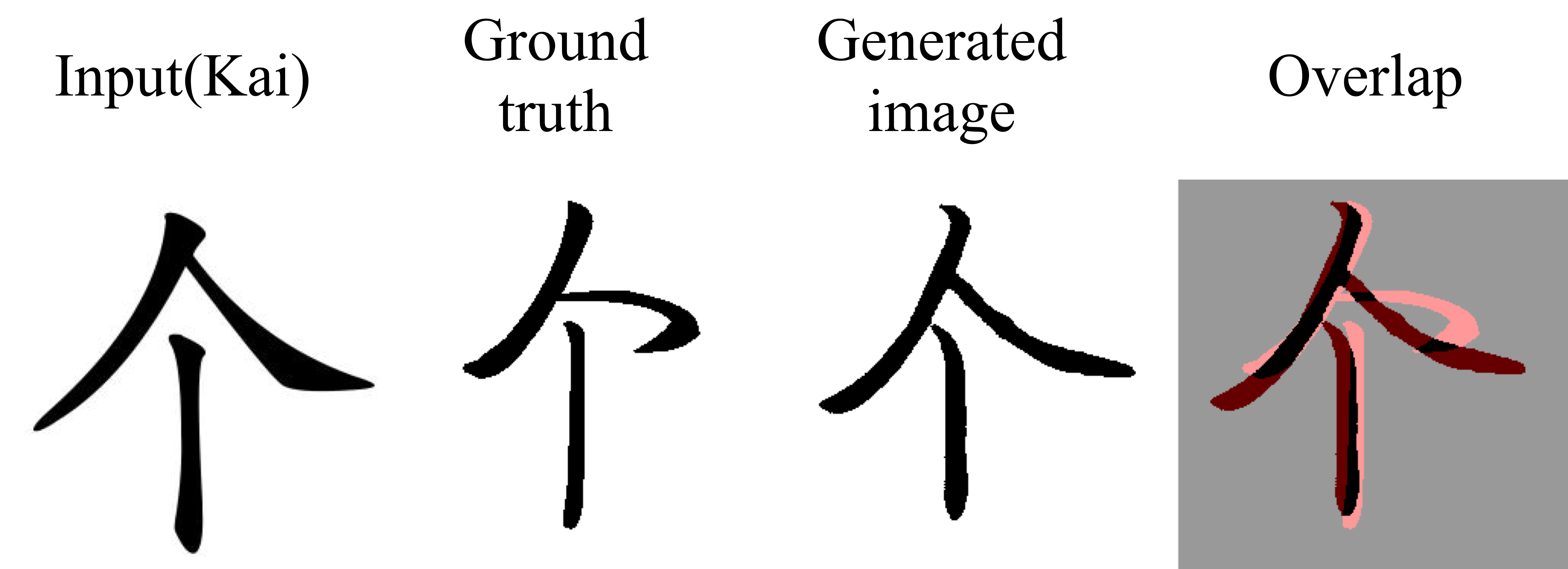} \label{fig: exceptge}}
        \subfloat[r][]{\includegraphics[width=0.3\textwidth]{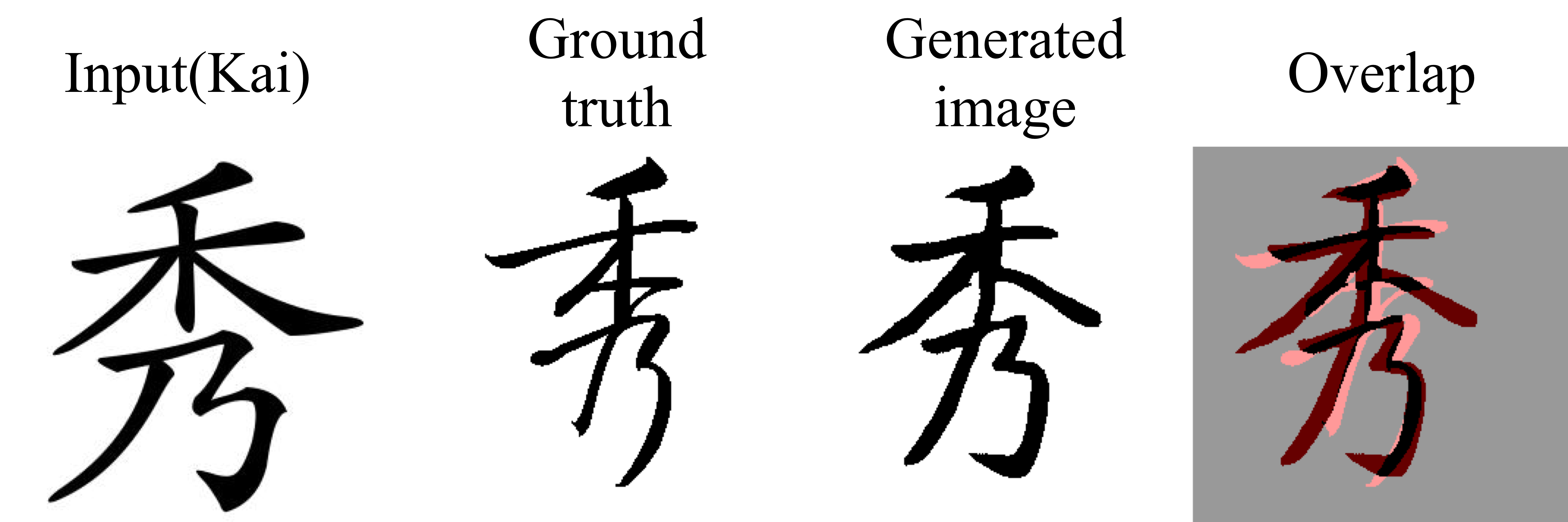} \label{fig: exceptxiu}}\hfill
        
        \subfloat[c][]{\includegraphics[width=0.25\textwidth]{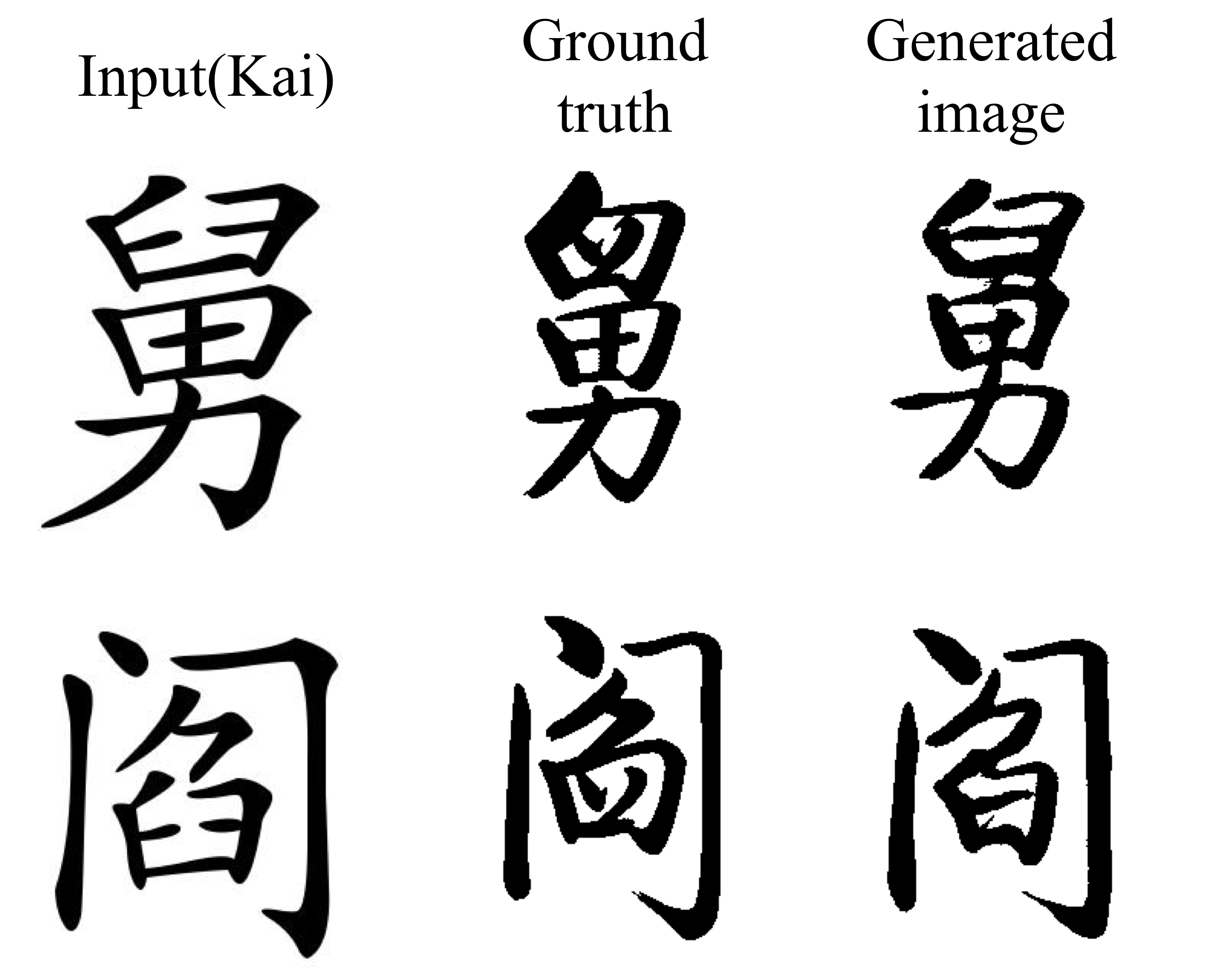} \label{fig: exceptjiuyan}}
        \subfloat[c][]{\includegraphics[width=0.25\textwidth]{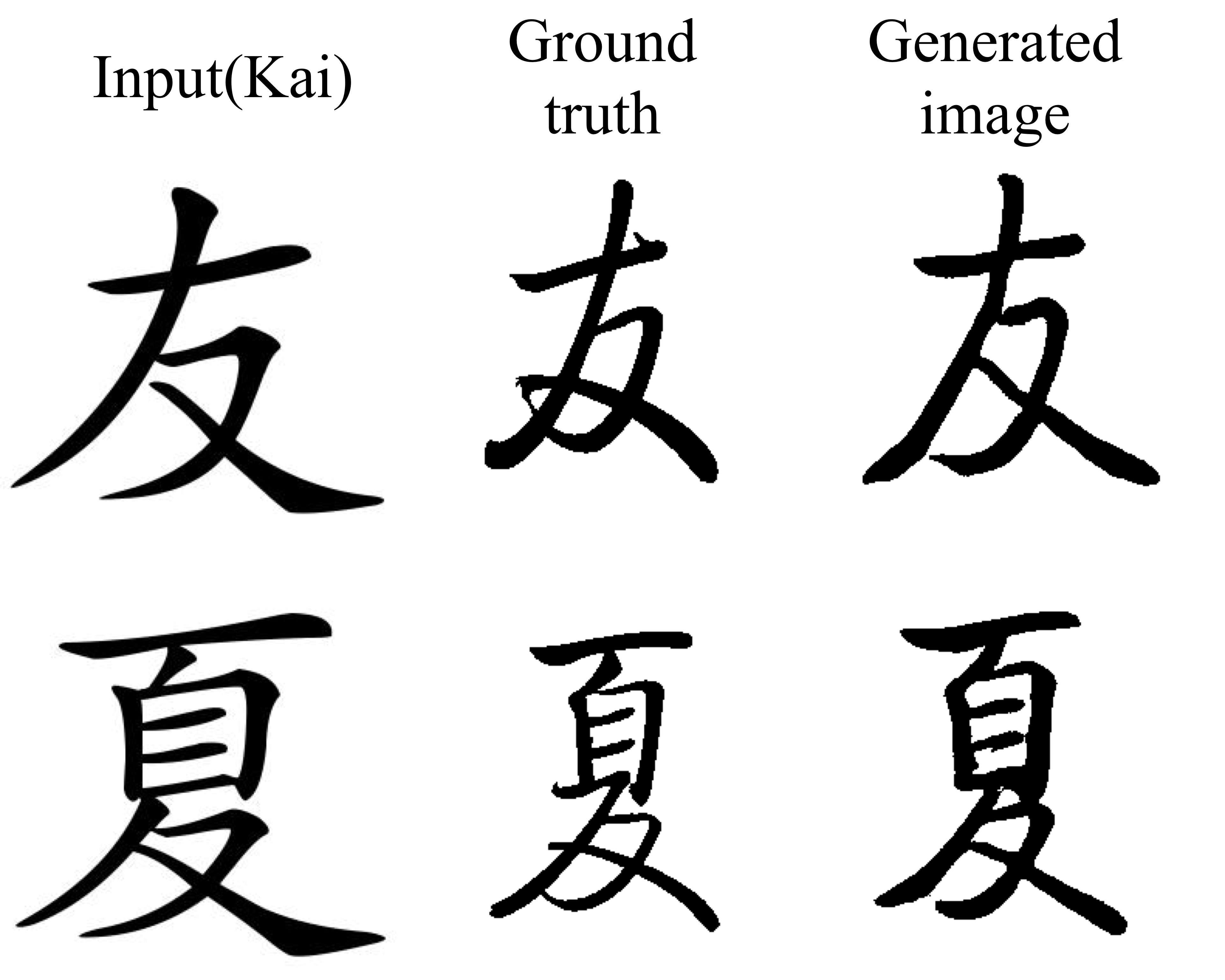} \label{fig: exceptyouxia}} 
        
        \caption{Samples different from the ground truth images.}
        \label{fig: exceptchars}
    \end{figure}

	Nevertheless there is a certain gap between our generation calligraphy images and ground truth images of QiGong font. There are significant differences between the generation images and the ground truth images of QiGong font when look at the details. Figure \ref{fig: exceptchars} shows some samples that the style is obviously different from the ground truth image. In those cases, the designers of QiGong font intentionally adopted those unique style of strokes and structures, which are obviously different from other characters. So it is difficult to learn these especial features by our approach. Therefore those generation images of QiGong font are different from the ground truth images.

\begin{figure}[!htbp]
	    \centering
	    \includegraphics[width=0.5\textwidth]{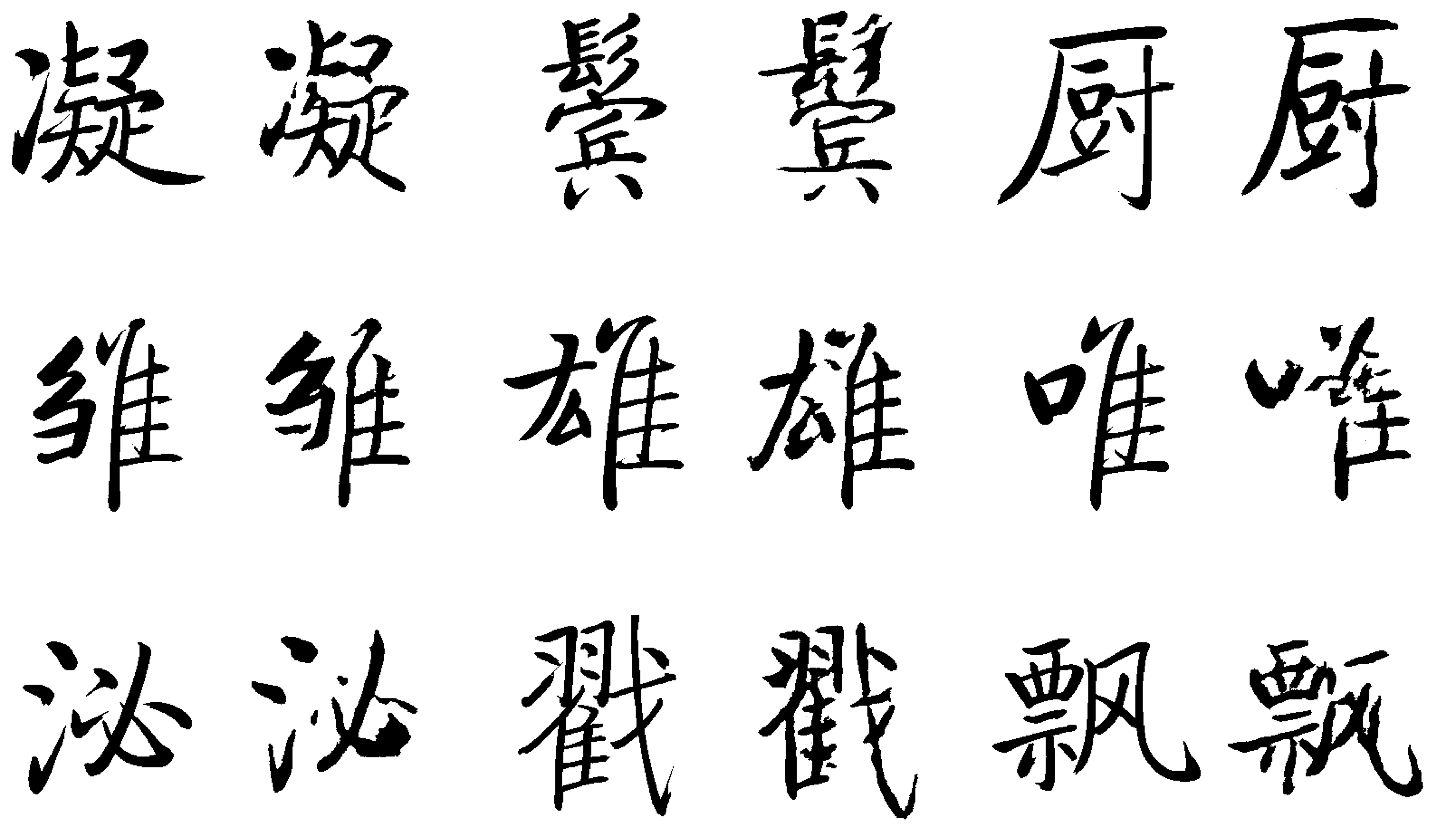}
	    \caption{Failed cases of generated results.}
	    \label{fig: faildsamples}
	\end{figure}

\noindent There are a lot of failed cases our model generation on training set 3000. Figure \ref{fig: faildsamples} shows some failed samples. In those failed samples, strokes or partial components of characters are broken, and the complex structures are blurring. One important reason is that the training samples are insufficient and the similar structure appears too little in training samples.

\section{Conclusion and Future Work}
This paper proposed a novel calligraphy generation method with deep generative adversarial networks. The main contribution of this paper demonstrates the possibility of complex Chinese calligraphy font generating. The key techniques in the proposed approach include: (1) To guarantee their excellent quality and adequate quantity, the datasets are created using a high-precision calligraphy synthesis technique; (2) The deep GAN model used in calligraphy generating is one of the best generative models currently; (3) Professional calligraphers are invited to conduct a Turing test on the experimental results to evaluate the gap between generative calligraphy and human art level. The tests evaluate the widely used calligraphy fonts for validation and evaluation. The outcomes and evaluation of our studies show that the proposed approach is successful in high-precision Chinese calligraphy generating. 

In the future, we will develop a set of post-processing techniques for improving the quality of strokes with the knowledge of Chinese calligraphic esthetics. For enhanced supervised learning, the calligraphy style necessitates knowledge description and formalization, comparative investigations with various models. The experimental design needs to be modified and utilize various robot platforms, i.e. multi-degree-of-freedom series robot and parallel robots \cite{xiao2018configuration}. The model advancement and experimental efficacy need to be assessed and tested from multiple perspectives considering the current evaluation metrics are insufficient.

\section*{Author Contributions}
Conceptualization: Zhiguo Gong proposed the initial idea. Xiaoming Wang contributed further ideas; Methodology: Xiaoming Wang and Zhiguo Gong contributed with ideas and discussion of critical points; Conducting experiments: Xiaoming Wang; Writing - original draft preparation: Xiaoming Wang; Writing - review and editing: Zhiguo Gong, Xiaoming Wang; Supervision: Zhiguo Gong.

\section*{Ethical and Informed Consent for Data Used}
Informed consent was obtained from all the participants prior to the publication of this study.

\section*{Data Availability Statements}
The data that support the findings of this study are available from the corresponding author upon reasonable request.

\section*{Conflict of interests}
The authors declare no conflict of interests.

\section*{Acknowledgment}
This work was supported by the National Key Research and Development Program of China (2019YFB1600704), The Science and Technology Development Fund, Macau SAR (0068/2020/AGJ, SKL-IOTSC(UM)-2021-2023),  GDST (2020B1212030003), MYRG2022-00192-FST.

\begin{appendices}






\end{appendices}

\bibliography{reference.bib}


\begin{thebibliography}{28}
\ifx \bisbn   \undefined \def \bisbn  #1{ISBN #1}\fi
\ifx \binits  \undefined \def \binits#1{#1}\fi
\ifx \bauthor  \undefined \def \bauthor#1{#1}\fi
\ifx \batitle  \undefined \def \batitle#1{#1}\fi
\ifx \bjtitle  \undefined \def \bjtitle#1{#1}\fi
\ifx \bvolume  \undefined \def \bvolume#1{\textbf{#1}}\fi
\ifx \byear  \undefined \def \byear#1{#1}\fi
\ifx \bissue  \undefined \def \bissue#1{#1}\fi
\ifx \bfpage  \undefined \def \bfpage#1{#1}\fi
\ifx \blpage  \undefined \def \blpage #1{#1}\fi
\ifx \burl  \undefined \def \burl#1{\textsf{#1}}\fi
\ifx \doiurl  \undefined \def \doiurl#1{\url{https://doi.org/#1}}\fi
\ifx \betal  \undefined \def \betal{\textit{et al.}}\fi
\ifx \binstitute  \undefined \def \binstitute#1{#1}\fi
\ifx \binstitutionaled  \undefined \def \binstitutionaled#1{#1}\fi
\ifx \bctitle  \undefined \def \bctitle#1{#1}\fi
\ifx \beditor  \undefined \def \beditor#1{#1}\fi
\ifx \bpublisher  \undefined \def \bpublisher#1{#1}\fi
\ifx \bbtitle  \undefined \def \bbtitle#1{#1}\fi
\ifx \bedition  \undefined \def \bedition#1{#1}\fi
\ifx \bseriesno  \undefined \def \bseriesno#1{#1}\fi
\ifx \blocation  \undefined \def \blocation#1{#1}\fi
\ifx \bsertitle  \undefined \def \bsertitle#1{#1}\fi
\ifx \bsnm \undefined \def \bsnm#1{#1}\fi
\ifx \bsuffix \undefined \def \bsuffix#1{#1}\fi
\ifx \bparticle \undefined \def \bparticle#1{#1}\fi
\ifx \barticle \undefined \def \barticle#1{#1}\fi
\bibcommenthead
\ifx \bconfdate \undefined \def \bconfdate #1{#1}\fi
\ifx \botherref \undefined \def \botherref #1{#1}\fi
\ifx \url \undefined \def \url#1{\textsf{#1}}\fi
\ifx \bchapter \undefined \def \bchapter#1{#1}\fi
\ifx \bbook \undefined \def \bbook#1{#1}\fi
\ifx \bcomment \undefined \def \bcomment#1{#1}\fi
\ifx \oauthor \undefined \def \oauthor#1{#1}\fi
\ifx \citeauthoryear \undefined \def \citeauthoryear#1{#1}\fi
\ifx \endbibitem  \undefined \def \endbibitem {}\fi
\ifx \bconflocation  \undefined \def \bconflocation#1{#1}\fi
\ifx \arxivurl  \undefined \def \arxivurl#1{\textsf{#1}}\fi
\csname PreBibitemsHook\endcsname

\bibitem{kobayashi2022generative}
\begin{barticle}
\bauthor{\bsnm{Kobayashi}, \binits{R.}},
\bauthor{\bsnm{Katsura}, \binits{S.}}:
\batitle{A generative model of calligraphy based on image and human motion}.
\bjtitle{Precision Engineering}
\bvolume{77},
\bfpage{340}--\blpage{348}
(\byear{2022})
\end{barticle}
\endbibitem

\bibitem{yang2014feature}
\begin{bchapter}
\bauthor{\bsnm{Yang}, \binits{L.-J.}},
\bauthor{\bsnm{Xu}, \binits{T.-C.}},
\bauthor{\bsnm{Li}, \binits{X.-S.}},
\bauthor{\bsnm{Wu}, \binits{E.-H.}}:
\bctitle{Feature-oriented writing process reproduction of chinese calligraphic artwork}.
In: \bbtitle{SIGGRAPH Asia 2014 Technical Briefs},
pp. \bfpage{1}--\blpage{4}
(\byear{2014})
\end{bchapter}
\endbibitem

\bibitem{wong2008model}
\begin{barticle}
\bauthor{\bsnm{Wong}, \binits{S.T.}},
\bauthor{\bsnm{Leung}, \binits{H.}},
\bauthor{\bsnm{Ip}, \binits{H.H.}}:
\batitle{Model-based analysis of chinese calligraphy images}.
\bjtitle{Computer Vision and Image Understanding}
\bvolume{109}(\bissue{1}),
\bfpage{69}--\blpage{85}
(\byear{2008})
\end{barticle}
\endbibitem

\bibitem{chen2018gated}
\begin{barticle}
\bauthor{\bsnm{Chen}, \binits{X.}},
\bauthor{\bsnm{Xu}, \binits{C.}},
\bauthor{\bsnm{Yang}, \binits{X.}},
\bauthor{\bsnm{Song}, \binits{L.}},
\bauthor{\bsnm{Tao}, \binits{D.}}:
\batitle{Gated-gan: Adversarial gated networks for multi-collection style transfer}.
\bjtitle{IEEE Transactions on Image Processing}
\bvolume{28}(\bissue{2}),
\bfpage{546}--\blpage{560}
(\byear{2018})
\end{barticle}
\endbibitem

\bibitem{wang2020deep}
\begin{barticle}
\bauthor{\bsnm{Wang}, \binits{W.-Y.}},
\bauthor{\bsnm{Hsu}, \binits{M.-J.}},
\bauthor{\bsnm{Yu}, \binits{L.-A.}},
\bauthor{\bsnm{Chien}, \binits{Y.-H.}},
\bauthor{\bsnm{Hsu}, \binits{C.-C.}}:
\batitle{Deep learning-based hypothesis generation model and its application on virtual chinese calligraphy-writing robot}.
\bjtitle{IEEE Access}
\bvolume{8},
\bfpage{87243}--\blpage{87251}
(\byear{2020})
\end{barticle}
\endbibitem

\bibitem{liang2020robot}
\begin{barticle}
\bauthor{\bsnm{Liang}, \binits{D.-t.}},
\bauthor{\bsnm{Liang}, \binits{D.}},
\bauthor{\bsnm{Xing}, \binits{S.-m.}},
\bauthor{\bsnm{Li}, \binits{P.}},
\bauthor{\bsnm{Wu}, \binits{X.-c.}}:
\batitle{A robot calligraphy writing method based on style transferring algorithm and similarity evaluation}.
\bjtitle{Intelligent Service Robotics}
\bvolume{13},
\bfpage{137}--\blpage{146}
(\byear{2020})
\end{barticle}
\endbibitem

\bibitem{jian2019learning}
\begin{barticle}
\bauthor{\bsnm{Jian}, \binits{M.}},
\bauthor{\bsnm{Dong}, \binits{J.}},
\bauthor{\bsnm{Gong}, \binits{M.}},
\bauthor{\bsnm{Yu}, \binits{H.}},
\bauthor{\bsnm{Nie}, \binits{L.}},
\bauthor{\bsnm{Yin}, \binits{Y.}},
\bauthor{\bsnm{Lam}, \binits{K.-M.}}:
\batitle{Learning the traditional art of chinese calligraphy via three-dimensional reconstruction and assessment}.
\bjtitle{IEEE Transactions on Multimedia}
\bvolume{22}(\bissue{4}),
\bfpage{970}--\blpage{979}
(\byear{2019})
\end{barticle}
\endbibitem

\bibitem{chao2017robot}
\begin{barticle}
\bauthor{\bsnm{Chao}, \binits{F.}},
\bauthor{\bsnm{Huang}, \binits{Y.}},
\bauthor{\bsnm{Zhang}, \binits{X.}},
\bauthor{\bsnm{Shang}, \binits{C.}},
\bauthor{\bsnm{Yang}, \binits{L.}},
\bauthor{\bsnm{Zhou}, \binits{C.}},
\bauthor{\bsnm{Hu}, \binits{H.}},
\bauthor{\bsnm{Lin}, \binits{C.-M.}}:
\batitle{A robot calligraphy system: From simple to complex writing by human gestures}.
\bjtitle{Engineering Applications of Artificial Intelligence}
\bvolume{59},
\bfpage{1}--\blpage{14}
(\byear{2017})
\end{barticle}
\endbibitem

\bibitem{chao2018use}
\begin{barticle}
\bauthor{\bsnm{Chao}, \binits{F.}},
\bauthor{\bsnm{Huang}, \binits{Y.}},
\bauthor{\bsnm{Lin}, \binits{C.-M.}},
\bauthor{\bsnm{Yang}, \binits{L.}},
\bauthor{\bsnm{Hu}, \binits{H.}},
\bauthor{\bsnm{Zhou}, \binits{C.}}:
\batitle{Use of automatic chinese character decomposition and human gestures for chinese calligraphy robots}.
\bjtitle{IEEE Transactions on Human-Machine Systems}
\bvolume{49}(\bissue{1}),
\bfpage{47}--\blpage{58}
(\byear{2018})
\end{barticle}
\endbibitem

\bibitem{ma2016aesthetics}
\begin{barticle}
\bauthor{\bsnm{Ma}, \binits{Z.}},
\bauthor{\bsnm{Su}, \binits{J.}}:
\batitle{Aesthetics evaluation for robotic chinese calligraphy}.
\bjtitle{IEEE Transactions on Cognitive and Developmental Systems}
\bvolume{9}(\bissue{1}),
\bfpage{80}--\blpage{90}
(\byear{2016})
\end{barticle}
\endbibitem

\bibitem{wu2023internal}
\begin{botherref}
\oauthor{\bsnm{Wu}, \binits{R.}},
\oauthor{\bsnm{Chao}, \binits{F.}},
\oauthor{\bsnm{Zhou}, \binits{C.}},
\oauthor{\bsnm{Chang}, \binits{X.}},
\oauthor{\bsnm{Yang}, \binits{L.}},
\oauthor{\bsnm{Shang}, \binits{C.}},
\oauthor{\bsnm{Zhang}, \binits{Z.}},
\oauthor{\bsnm{Shen}, \binits{Q.}}:
Internal model control structure inspired robotic calligraphy system.
IEEE Transactions on Industrial Informatics
(2023)
\end{botherref}
\endbibitem

\bibitem{gao2019data}
\begin{barticle}
\bauthor{\bsnm{Gao}, \binits{X.}},
\bauthor{\bsnm{Zhou}, \binits{C.}},
\bauthor{\bsnm{Chao}, \binits{F.}},
\bauthor{\bsnm{Yang}, \binits{L.}},
\bauthor{\bsnm{Lin}, \binits{C.-M.}},
\bauthor{\bsnm{Xu}, \binits{T.}},
\bauthor{\bsnm{Shang}, \binits{C.}},
\bauthor{\bsnm{Shen}, \binits{Q.}}:
\batitle{A data-driven robotic chinese calligraphy system using convolutional auto-encoder and differential evolution}.
\bjtitle{Knowledge-Based Systems}
\bvolume{182},
\bfpage{104802}
(\byear{2019})
\end{barticle}
\endbibitem

\bibitem{GUO2022Calligraphy}
\begin{barticle}
\bauthor{\bsnm{Dong-mei}, \binits{G.}},
\bauthor{\bsnm{Hua-song}, \binits{M.}}:
\batitle{Survey of calligraphy robot}.
\bjtitle{Control and Decision}
\bvolume{37}(\bissue{7}),
\bfpage{1665}--\blpage{1674}
(\byear{2022})
\end{barticle}
\endbibitem

\bibitem{zhang2019survey}
\begin{barticle}
\bauthor{\bsnm{Zhang}, \binits{J.}}:
\batitle{A survey of digital calligraphy.}
\bjtitle{SCIENTIA SINICA Informationis}
\bvolume{49},
\bfpage{143}--\blpage{158}
(\byear{2019})
\end{barticle}
\endbibitem

\bibitem{DBLP:journals/expert/XuJJLP09}
\begin{barticle}
\bauthor{\bsnm{Xu}, \binits{S.}},
\bauthor{\bsnm{Jiang}, \binits{H.}},
\bauthor{\bsnm{Jin}, \binits{T.}},
\bauthor{\bsnm{Lau}, \binits{F.C.M.}},
\bauthor{\bsnm{Pan}, \binits{Y.}}:
\batitle{Automatic generation of chinese calligraphic writings with style imitation}.
\bjtitle{{IEEE} Intelligent Systems}
\bvolume{24}(\bissue{2}),
\bfpage{44}--\blpage{53}
(\byear{2009}).
\doiurl{10.1109/MIS.2009.23}
\end{barticle}
\endbibitem

\bibitem{DBLP:journals/expert/DongXZGP08}
\begin{barticle}
\bauthor{\bsnm{Dong}, \binits{J.}},
\bauthor{\bsnm{Xu}, \binits{M.}},
\bauthor{\bsnm{Zhang}, \binits{X.}},
\bauthor{\bsnm{Gao}, \binits{Y.}},
\bauthor{\bsnm{Pan}, \binits{Y.}}:
\batitle{The creation process of chinese calligraphy and emulation of imagery thinking}.
\bjtitle{{IEEE} Intelligent Systems}
\bvolume{23}(\bissue{6}),
\bfpage{56}--\blpage{62}
(\byear{2008}).
\doiurl{10.1109/MIS.2008.110}
\end{barticle}
\endbibitem

\bibitem{Zong:2014:SAP:2892753.2892971}
\begin{bchapter}
\bauthor{\bsnm{Zong}, \binits{A.}},
\bauthor{\bsnm{Zhu}, \binits{Y.}}:
\bctitle{Strokebank: Automating personalized chinese handwriting generation}.
In: \bbtitle{Proceedings of the Twenty-Eighth AAAI Conference on Artificial Intelligence}.
\bsertitle{AAAI'14},
pp. \bfpage{3024}--\blpage{3029}.
\bpublisher{AAAI Press}, \blocation{???}
(\byear{2014}).
\burl{http://dl.acm.org/citation.cfm?id=2892753.2892971}
\end{bchapter}
\endbibitem

\bibitem{DBLP:journals/corr/ChangG17}
\begin{botherref}
Chinese typography transfer.
CoRR
\textbf{abs/1707.04904}
(2017)
{\href{https://arxiv.org/abs/1707.04904}{{arXiv:1707.04904}}}.
Withdrawn.
\end{botherref}
\endbibitem

\bibitem{Lian:2016:AGL:3005358.3005371}
\begin{bchapter}
\bauthor{\bsnm{Lian}, \binits{Z.}},
\bauthor{\bsnm{Zhao}, \binits{B.}},
\bauthor{\bsnm{Xiao}, \binits{J.}}:
\bctitle{Automatic generation of large-scale handwriting fonts via style learning}.
In: \bbtitle{SIGGRAPH ASIA 2016 Technical Briefs}.
\bsertitle{SA '16},
pp. \bfpage{12}--\blpage{1124}.
\bpublisher{ACM},
\blocation{New York, NY, USA}
(\byear{2016}).
\doiurl{10.1145/3005358.3005371}.
\burl{http://doi.acm.org/10.1145/3005358.3005371}
\end{bchapter}
\endbibitem

\bibitem{Auto-En17}
\begin{botherref}
\oauthor{\bsnm{Lyu}, \binits{P.}},
\oauthor{\bsnm{Bai}, \binits{X.}},
\oauthor{\bsnm{Yao}, \binits{C.}},
\oauthor{\bsnm{Zhu}, \binits{Z.}},
\oauthor{\bsnm{Huang}, \binits{T.}},
\oauthor{\bsnm{Liu}, \binits{W.}}:
Auto-encoder guided {GAN} for chinese calligraphy synthesis.
CoRR
\textbf{abs/1706.08789}
(2017)
{\href{https://arxiv.org/abs/1706.08789}{{arXiv:1706.08789}}}
\end{botherref}
\endbibitem

\bibitem{wang2023generative}
\begin{botherref}
\oauthor{\bsnm{Wang}, \binits{X.}},
\oauthor{\bsnm{Yang}, \binits{Y.}},
\oauthor{\bsnm{Wang}, \binits{W.}},
\oauthor{\bsnm{Zhou}, \binits{Y.}},
\oauthor{\bsnm{Yin}, \binits{Y.}},
\oauthor{\bsnm{Gong}, \binits{Z.}}:
Generative adversarial networks based motion learning towards robotic calligraphy synthesis.
CAAI Transactions on Intelligence Technology
(2023)
\end{botherref}
\endbibitem

\bibitem{goodfellow2014generative}
\begin{botherref}
\oauthor{\bsnm{Goodfellow}, \binits{I.}},
\oauthor{\bsnm{Pouget-Abadie}, \binits{J.}},
\oauthor{\bsnm{Mirza}, \binits{M.}},
\oauthor{\bsnm{Xu}, \binits{B.}},
\oauthor{\bsnm{Warde-Farley}, \binits{D.}},
\oauthor{\bsnm{Ozair}, \binits{S.}},
\oauthor{\bsnm{Courville}, \binits{A.}},
\oauthor{\bsnm{Bengio}, \binits{Y.}}:
Generative adversarial nets.
Advances in neural information processing systems
\textbf{27}
(2014)
\end{botherref}
\endbibitem

\bibitem{cgans}
\begin{botherref}
\oauthor{\bsnm{Mirza}, \binits{M.}},
\oauthor{\bsnm{Osindero}, \binits{S.}}:
Conditional generative adversarial nets.
CoRR
\textbf{abs/1411.1784}
(2014)
{\href{https://arxiv.org/abs/1411.1784}{{arXiv:1411.1784}}}
\end{botherref}
\endbibitem

\bibitem{pix2pix}
\begin{botherref}
\oauthor{\bsnm{Isola}, \binits{P.}},
\oauthor{\bsnm{Zhu}, \binits{J.}},
\oauthor{\bsnm{Zhou}, \binits{T.}},
\oauthor{\bsnm{Efros}, \binits{A.A.}}:
Image-to-image translation with conditional adversarial networks.
CoRR
\textbf{abs/1611.07004}
(2016)
{\href{https://arxiv.org/abs/1611.07004}{{arXiv:1611.07004}}}
\end{botherref}
\endbibitem

\bibitem{crossdomain}
\begin{botherref}
\oauthor{\bsnm{Taigman}, \binits{Y.}},
\oauthor{\bsnm{Polyak}, \binits{A.}},
\oauthor{\bsnm{Wolf}, \binits{L.}}:
Unsupervised cross-domain image generation.
CoRR
\textbf{abs/1611.02200}
(2016)
{\href{https://arxiv.org/abs/1611.02200}{{arXiv:1611.02200}}}
\end{botherref}
\endbibitem

\bibitem{ssim}
\begin{barticle}
\bauthor{\bsnm{Wang}, \binits{Z.}},
\bauthor{\bsnm{Bovik}, \binits{A.C.}},
\bauthor{\bsnm{Sheikh}, \binits{H.R.}},
\bauthor{\bsnm{Simoncelli}, \binits{E.P.}}:
\batitle{Image quality assessment: from error visibility to structural similarity}.
\bjtitle{IEEE Transactions on Image Processing}
\bvolume{13}(\bissue{4}),
\bfpage{600}--\blpage{612}
(\byear{2004}).
\doiurl{10.1109/TIP.2003.819861}
\end{barticle}
\endbibitem

\bibitem{adam}
\begin{botherref}
\oauthor{\bsnm{Kingma}, \binits{D.P.}},
\oauthor{\bsnm{Ba}, \binits{J.}}:
Adam: {A} method for stochastic optimization.
CoRR
\textbf{abs/1412.6980}
(2014)
{\href{https://arxiv.org/abs/1412.6980}{{arXiv:1412.6980}}}
\end{botherref}
\endbibitem

\bibitem{xiao2018configuration}
\begin{barticle}
\bauthor{\bsnm{Xiao}, \binits{X.}},
\bauthor{\bsnm{Li}, \binits{Y.}},
\bauthor{\bsnm{Wang}, \binits{X.}}:
\batitle{Configuration analysis and design of a multidimensional tele-operator based on a 3-p (4s) parallel mechanism}.
\bjtitle{Journal of Intelligent \& Robotic Systems}
\bvolume{90},
\bfpage{339}--\blpage{348}
(\byear{2018})
\end{barticle}
\endbibitem

\end{thebibliography}


\end{document}